\documentclass[]{interact}

\usepackage{epstopdf}
\usepackage[caption=false]{subfig}
\usepackage[doublespacing]{setspace}

\usepackage[natbibapa,nodoi]{apacite}
\renewcommand\bibliographytypesize{\fontsize{10}{12}\selectfont}

\usepackage{lineno}
\usepackage{amsmath}

\usepackage{algorithm}
\usepackage{algpseudocode}
\usepackage{array}
\usepackage{graphicx}
\usepackage{url}
\usepackage{booktabs}
\usepackage{multirow}
\usepackage{bbm}  
\usepackage{float}
\usepackage{xcolor}
\usepackage{comment}
\usepackage{algpseudocode}
\usepackage{mathtools}

\theoremstyle{plain}

\theoremstyle{definition}

\theoremstyle{remark}

\usepackage{geometry}
\begin{document}

\articletype{International Journal of Geographical Information Science}

\title{STICC: A multivariate spatial clustering method for repeated geographic pattern discovery with consideration of spatial contiguity}


\author{
\name{Yuhao Kang\textsuperscript{a}, Kunlin Wu\textsuperscript{b}, Song Gao\textsuperscript{a*}\thanks{*Corresponding Author: Song Gao. Email: song.gao@wisc.edu}, Ignavier Ng\textsuperscript{c}, Jinmeng Rao\textsuperscript{a}, Shan Ye\textsuperscript{d}, Fan Zhang\textsuperscript{e} and Teng Fei\textsuperscript{b}}
\affil{
\textsuperscript{a} GeoDS Lab, Department of Geography, University of Wisconsin-Madison, WI, United States;\\
\textsuperscript{b} School of Resources and Environmental Sciences, Wuhan University, Wuhan, China;\\
\textsuperscript{c} Department of Philosophy, Carnegie Mellon University, PA, United States;\\
\textsuperscript{d} Department of Geoscience, University of Wisconsin-Madison, WI, United States;\\
\textsuperscript{e} Senseable City Lab, Massachusetts Institute of Technology, MA, United States.
}
}

\maketitle

\begin{abstract}
Spatial clustering has been widely used for spatial data mining and knowledge discovery. An ideal multivariate spatial clustering should consider both spatial contiguity and aspatial attributes. Existing spatial clustering approaches may face challenges for discovering repeated geographic patterns with spatial contiguity maintained. In this paper, we propose a Spatial Toeplitz Inverse Covariance-Based Clustering (STICC) method that considers both attributes and spatial relationships of geographic objects for multivariate spatial clustering. A subregion is created for each geographic object serving as the basic unit when performing clustering. A Markov random field is then constructed to characterize the attribute dependencies of subregions. Using a spatial consistency strategy, nearby objects are encouraged to belong to the same cluster. To test the performance of the proposed STICC algorithm, we apply it in two use cases. The comparison results with several baseline methods show that the STICC outperforms others significantly in terms of adjusted rand index and macro-F1 score.
Join count statistics is also calculated and shows that the spatial contiguity is well preserved by STICC. Such a spatial clustering method may benefit various applications in the fields of geography, remote sensing, transportation, and urban planning, etc.
\end{abstract}


\section{Introduction}\label{sec:intro}
A place typically has multiple features, such as environmental and socioeconomic variables.
The spatial distribution of similar places may have the following two scenarios.
On the one hand, as stated by the \textit{first law of geography} that ``near things are more related than distant things," nearby places share similar characteristics \citep{tobler1970computer,goodchild2004giscience, zhu2018spatial}.
For example, due to the nature of spatial dependence in geographic phenomena, two nearby meteorological stations may observe similar temperature, precipitation and humidity;
two adjacent neighborhoods may have the same urban functions (e.g., residential area, commercial area, or educational area), because their socioeconomic characteristics are highly correlated \citep{yuan2014discovering, gao2017extracting,xing2018integrating}. 
On the other hand, some places located in different areas may have similar attributes.
For instance, Italy and California, US, have the same Mediterranean climate type; airports in two different cities are both transportation hubs.
Hence, places that are not spatially adjacent to each other may still belong to the same group since their attributes are quite similar.
This phenomenon, there exist places with similar attributes that are either nearby or far away, can be frequently seen on the earth.

Uncovering such patterns, which can be named as a repeated geographic pattern discovery (RGPD) problem, i.e., finding out \textit{repeated} groups of similar places across space and maintaining the \textit{spatial contiguity} of geographic patterns within each subcluster, requires multivariate spatial clustering \citep{murray1998cluster,miller2009geographic}.
Spatial clustering aims at partitioning spatial data into a series of meaningful subgroups, and has played important roles in spatio-temporal data mining and knowledge discovery \citep{duque2007supervised,aldstadt2010spatial,liu2012density}.
By identifying spatial clusters, geographic objects with similar attributes or adjacent locations are grouped into same clusters, and are dissimilar or distant from other clusters.
Detecting these clusters is necessary for a series of spatial analyses and GIS applications such as land use classification, cartographic generalization, public health, and soil mapping \citep{liqiang2013spatial,esri2021spatial, wang2020public}.

\begin{figure}[h]
    \centering
    \includegraphics[width=\textwidth]{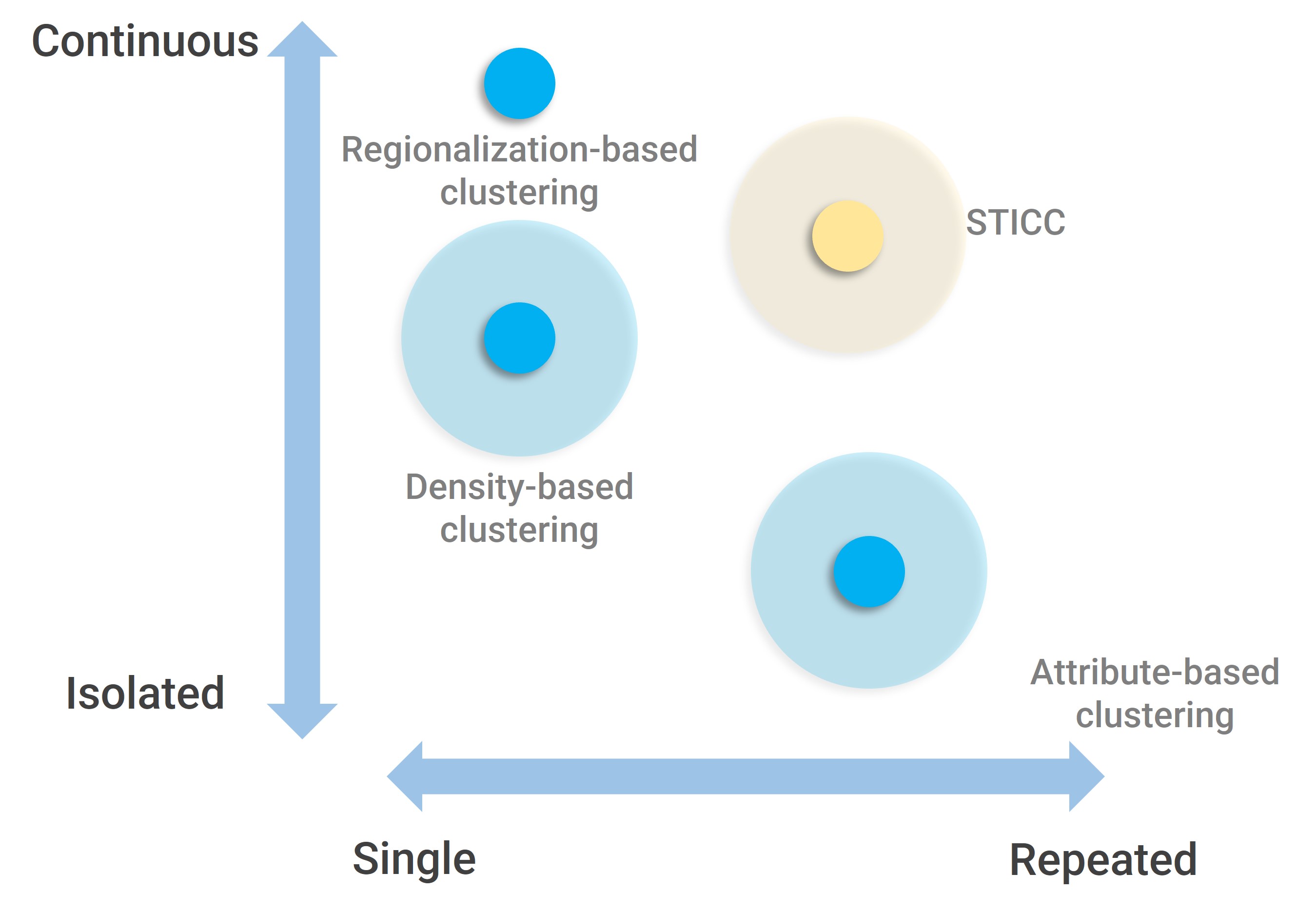}
    \caption{
    Characteristics of geographic phenomena and corresponding spatial clustering methods: horizontal-axis represents repeated patterns, and vertical-axis illustrates the degree of spatial contiguity. The translucent area represent the uncertainty of locations for clustering algorithms.}
    \label{fig:framework}
\end{figure}

\begin{figure}[h]
    \centering
    \includegraphics[width=\textwidth]{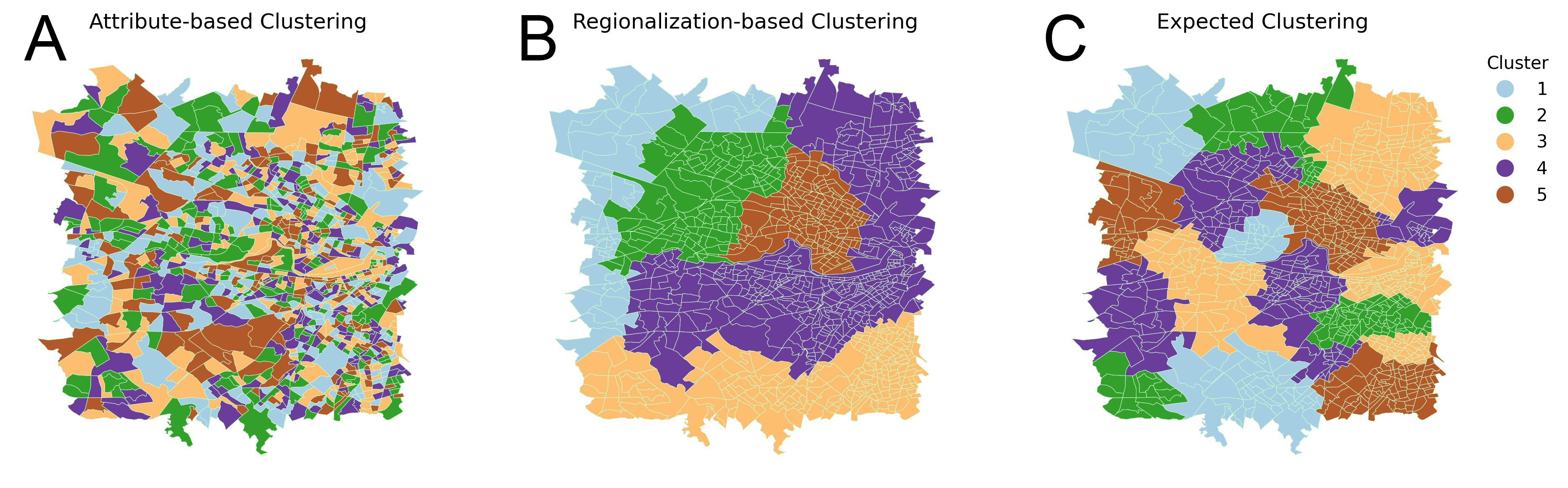}
    \caption{
    Example maps of spatial clustering results based on a synthetic dataset: (A) attributed-based clustering approaches; (B) regionalization-based clustering approaches; (C) expected clustering approaches (e.g., the proposed STICC) for RGPD problem. There are five clusters of regions in the study area expressed in different colors.}
    \label{fig:example_map}
\end{figure}

In fact, most geographic phenomena have the following two dimensions of properties, repeated patterns in attributes (single vs. repeated), and spatial contiguity (isolated vs. continuous).
In this paper, \textit{repeated patterns} refer to whether regions with similar geographic phenomena/attributes (i.e., belong to the same cluster) could appear in different locations;
and the purpose of \textit{spatial contiguity} is to assess whether and to what degree the attributes of geographic objects are spatially dependent.
It should be noted that, for polygons, nearby geographic objects are their adjacent neighbors with shared boundary; while for points, Delaunay triangulation or other types of connectivity among points might be constructed first for obtaining their neighboring points.
Existing spatial clustering methods can be divided into the following three categories: attribute-based clustering, regionalization-based clustering, and density-based clustering, though there are several other taxonomies \citep{deng2011adaptive}. 
These methods are suitable for single and continuous, or repeated and isolated geographic pattern discovery. 
However, these methods may face challenges in handling RGPD problem, i.e., discovering repeated patterns with spatial contiguity preserved.
Ideally, a multivariate spatial clustering for solving the RGPD problem should consider both spatial and aspatial attributes, that is, (1) geographic objects with similar attributes are grouped together and such groups may occur repeatedly across the space, (2) and nearby objects in physical space are encouraged to be assigned in the same cluster to maintain spatial contiguity.
The two-dimensional characteristics of geographic phenomena (repeated patterns in attributes and spatial contiguity) as well as their corresponding spatial clustering methods are plotted in Figure \ref{fig:framework}. 
In the following paragraphs, we describe more details for each category of spatial clustering methods.
We also produce several example maps, as shown in Figure \ref{fig:example_map}, based on a synthetic dataset to help demonstrate the characteristics of potential outputs by using different spatial clustering methods. 

Attribute-based clustering methods, by definition, group geographic objects according to their multiple attributes.
One way to perform such kind of clustering analysis is solely based on attributes while ignoring spatial relationships, such as \textit{K-Means} \citep{macqueen1967some}, \textit{BIRCH} \citep{zhang1996birch}, \textit{CURE} \citep{guha1998cure}, and \textit{SOM} (Self-organized Map) \citep{baccao2005self}.
The underlying hypothesis of these algorithms is that spatial dependence has been embodied by these multi-dimensional attributes and thereby the spatial structures can be discovered.
Another commonly used solution in practice is to treat spatial coordinates as two additional weighted attributes \citep{webster1972computer, murray2000integrating}.
It has two limitations \citep{perruchet1983constrained}.
First, the importance of coordinates relies on the nature of phenomenon.
It is hard to decide how spatial and aspatial attributes should be combined and weighted \citep{duque2007supervised}.
Second, spatial relationships, such as distant, nearby, and adjacent, which play key roles in the spatial analysis, are usually underestimated, as geography might not be a dimension in the multidimensional space \citep{henriques2009geosom}.
When performing attribute-based clustering, a potential output map is shown in Figure \ref{fig:example_map}(A).
Geographic objects belonging to the same cluster may be distributed across space if without spatial contiguity consideration.
For example, regions that belong to cluster 2 (in green) may appear in multiple locations repeatedly across the entire study area but the spatial contiguity has been destroyed.
Hence, as suggested in Figure \ref{fig:framework}, attribute-based clustering may discover repeated patterns of geographic phenomenon, whereas the spatial contiguity of geographic patterns among different parts of the cluster may not be well preserved.

Density-based clustering methods such as DBSCAN \citep{ester1996density}, OPTICS \citep{ankerst1999optics}, ENCLUE \citep{hinneburg1998efficient}, and ADCN \citep{mai2018adcn}, are able to find out densely located geographic patterns by examining the number of nearby geographic objects.
They have been widely used in various GIS applications such as hotspot detection \citep{pei2006new,chen2018hispatialcluster, kang2019extracting}, urban areas of interest discovery \citep{hu2015extracting,liu2020investigating}, and taxi route and trajectory classification \citep{pei2015density,deng2019density, liu2021snn_flow, moayedi2019evaluation}.
Clusters of arbitrary shape can be discovered and the number of clusters need not to be predefined in most cases.
Because density-based clustering methods usually rely on the geometric information rather than the attributes of objects (though several studies have attempted to incorporate the object attributes into these methods \citep{liu2012density}), they are rarely used in finding groups of places with similar attributes.
For each cluster, geographic objects gather together, and the cluster only appears once in space, while distant places with similar attributes cannot be allocated to the same cluster.
Even two distant clusters with similar attributes are detected correctly, they are identified as two distinct groups.
In addition, density-based clustering encounters difficulties in finding evenly distributed groups.
In short, density-based clustering may not discover repeated spatial clusters, though relatively high spatial contiguity is maintained for each cluster, as shown in Figure \ref{fig:framework}.
It should be noted that the clustering results using attribute-based clustering or density-based approaches may also be influenced by the degree of spatial dependence of the input geographic data.
The output may still preserve the spatial contiguity to a certain degree. 
Such uncertainties are indicated by the translucent area in Figure \ref{fig:framework}.

The last category is regionalization-based clustering methods.
Such methods are usually used for solving the p-regions problem, which is defined as the aggregation of $n$ small areas into $p$ geographically connected regions \citep{duque2011p}.
It can partition space into a set of clustered regions where geographic objects are spatially connected inside those regions.
Graph-based methods are commonly used for solving such a problem such as SKATER \citep{assunccao2006efficient, aydin2018skater} and AUTOCLUST \citep{aldstadt2010spatial}.
The aggregated regions determined by the regionalization-based methods are unique.
Only nearby regions are aggregated to the same cluster, while distant regions with similar attributes cannot be assigned into the same cluster.
As displayed in the example map in Figure \ref{fig:example_map}(B), regions  belonging to cluster 3 are spatially connected but only locate at the bottom of the map.
The position of regionalization-based methods is thereby determined as illustrated in the Figure \ref{fig:framework}.
Consequently, similar to density-based methods, spatial contiguity can be well-preserved but no repeated spatial clusters are discovered by regionalization-based methods.

In view of all of these challenges encountered in current spatial clustering methods, we aim to develop a clustering method that can consider both spatial and aspatial attributes of geographic objects.
Such a method is expected to find repeated geographic patterns and maintain spatial contiguity simultaneously as shown in Figure \ref{fig:framework}.
In this paper, we develop a new method that addresses these issues, which is named as \textit{Spatial Toeplitz Inverse Covariance-based Clustering (STICC)}.
The algorithm is developed to achieve the balance between multi-dimensional attributes and spatial contiguity of geographic objects with the following two characteristics.
First, different from other clustering methods that usually treat each geographic object individually, we build off the algorithm using Markov random field (MRF) \citep{rue2005gaussian,koller09probabilistic}, a powerful tool that models partial correlation between different variables, to portray the dependencies of multi-dimensional attributes within a specific cluster. It builds a \textit{subregion} between a geographic object with its nearby objects.
Second, a \textit{spatial consistency} strategy is used to encourage the nearby geographic objects to belong to the same cluster.
Such a method is inspired by the \textit{Toeplitz inverse covariance-based clustering (TICC)} method proposed by \citet{hallac2017toeplitz} which has been widely adopted in various multivariate time series clustering applications.
It should be acknowledged that different from the time series clustering problem, the challenges of spatial clustering approaches are unique.
A time-series datum only contains one-dimensional timestamp values and thereby is naturally represented as a linearly ordered sequence, while the position of each spatial object is expressed as two-dimensional coordinates and thus such a linearly ordered sequence does not naturally exist for geographic objects.

The contribution of this paper is three-fold:
\begin{enumerate}
    \item We develop a novel spatial clustering method that considers not only the attributes of an object but also the spatial contiguity for multivariate repeated geographic pattern discovery (RGPD).
    \item We validate the reliability and effectiveness of the proposed method through experiments on both synthetic examples and real-world applications.
    \item We use the join count statistics to measure the spatial dependence of the clustering result for evaluating the performance of different clustering methods.
\end{enumerate}

This article is structured as follows. 
In section 2, we present the core idea of STICC and offer key technical implementation details, including MRF, Toeplitz matrices, and Toeplitz graphical lasso.
In section 3, we apply the method to two case studies to validate its performance, including synthetic datasets and real-world classification scenarios. 
The results for different clustering methods are compared and analyzed. 
We make some discussions regarding hyperparameter selections of the proposed algorithm and its implications for GeoAI studies, and acknowledge limitations and opportunities for future work in section 4.
Finally, we summarize and draw conclusions in section 5.

\section{Method}
In this section, we first outline the preliminary and required notations. We then motivate and describe the formulation of the proposed clustering problem as a mixed combinatorial and continuous optimization problem, which involves different components such as MRF, spatial Toeplitz matrix, and a penalty term to enforce spatial contiguity. Lastly, we develop an expectation-maximization (EM)-style procedure to solve the optimization problem.

It is worth noting that some part of the proposed method is inspired by the TICC method developed by \citet{hallac2017toeplitz}. However, as described in Section \ref{sec:intro}, the technical challenges are different as the TICC method is based on time series data for which a linearly ordered sequence naturally exists for the one-dimensional timestamp values, while we focus on the spatial clustering problem for which a linearly ordered sequence does not exist for the geographic objects expressed as two-dimensional coordinates.

\subsection{Preliminary}
The multivariate spatial clustering problem can be formulated as follows: For a given study area, clustering $N$ geographic objects (such as points, polylines, and polygons) into $K$ groups. The $D$ dimensional attributes of the geographic objects are defined as
\begin{equation}
    x=
    \begin{bmatrix}
    x_1^T\\
    x_2^T\\
    \vdots\\
    x_N^T\\
    \end{bmatrix}
    =
    \begin{bmatrix}
    x_{1,1} & x_{1,2} & \ldots & x_{1,D} \\
    x_{2,1} & x_{2,2} & \ldots & x_{2,D} \\
    \vdots  & \vdots  & \vdots & \vdots  \\
    x_{N,1} & x_{N,2} & \ldots & x_{N,D} 
    \end{bmatrix},
\end{equation}
where $x_n\in \mathbb{R}^D$ is a vector that corresponds to the $n$th multivariate geographic object, and $x_{n,i}$ refers to the $i$th attribute of the $n$th geographic object.
Furthermore, each geographic object $x_n$ is associated with a pair of coordinates (centroids are used for polygons), denoted as $c_n=(c_{n, 1},c_{n, 2})$.

\begin{figure}[h]
    \centering
    \includegraphics[width=\textwidth]{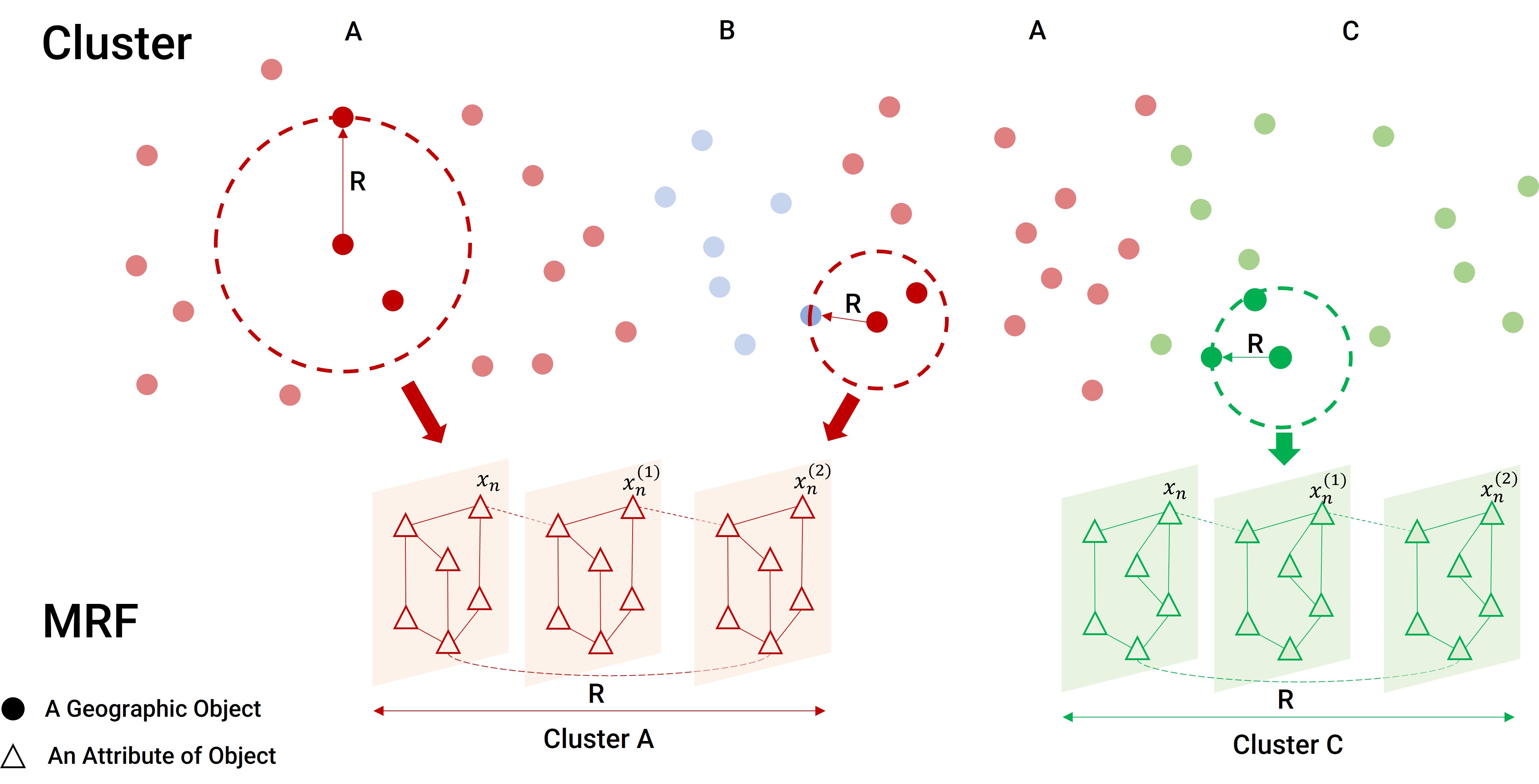}
    \caption{
    The idea of the proposed STICC method: geographic objects are grouped into clusters that are represented with different colors; a \textit{subregion} is built for each geographic object with its $R$-1 nearest neighbors (here subregion radius $R=3$); $x_n$ represents the center geographic object, and $x^{(r)}_n$ indicates its $r$th nearest neighbor; each cluster is characterized by an MRF to express (1) the interdependencies among attributes of different objects in the subregion (indicated by dashed lines), and (2) intradependencies among attributes of a single object (represented by solid lines); each attribute can be linked to all other attributes inside the MRF; each subregion is input into the optimization problem of STICC to learn the structure of MRF.}
    \label{fig:idea}
\end{figure}

Due to the existence of spatial dependence, nearby geographic objects may share similar characteristics and should be taken into account for spatial analysis.
Hence, we construct a \textit{subregion} of each point that includes several of its nearest neighbors.
All objects in subregions are fed into the model for clustering. 
Generally, there are two ways to define a subregion, using a search radius (symmetric) or k-nearest neighbors (asymmetric) \citep{mai2018adcn}. 
For asymmetric definition of neighborhood, the nearest neighbors of a geographic object might not always be symmetric.
For instance, a point $A$'s nearest neighbor might be $B$, and the nearest neighbor of $B$ might be another point $C$ while not $A$.
Here, we adopt the latter one as the number of objects in each subregion is fixed, and use the center geographic object as well as its nearest $(R-1)$ neighboring objects, that is a total of $R$ objects, for constructing the subregion. 
To avoid confusion, we use the term $R$ as the ``radius" of subregions that is equivalent to the number of geographic objects in each subregion, while we use $K$ as the number of clusters to be detected.
The nearest neighbors are determined according to the spatial matrix calculated based on the pairwise distance among objects.
For example, $R=1$ denotes that only the object itself is used to construct the subregion while the two nearest geographic objects of the center object are taken into account when $R=3$ as shown in Figure \ref{fig:idea}.

We then concatenate each object $x_n$ and its $R-1$ nearest objects $x_n^{(1)},\dots,x_n^{(R-1)}$, sorted in ascending order based on distance, into a vector; specifically, we write $X_n=(x_n, x_n^{(1)}, \dots, x_n^{(R-1)})\in\mathbb{R}^{DR}$ as the subregion in which $x_n$ is the centering geographic object.
In addition, denote by $X_n^{(1)}$ the nearest subregion of $X_n$.
The proposed STICC performs clustering on these stacked subregions but not on the set of all objects directly. This is different from the case of TICC that focuses on time series data, because a linearly ordered sequence naturally exists for time series and therefore the time window adopted by TICC is symmetric. 
The concept of time window for time series is analogous to our definition of subregions for spatial data, though the latter is not always symmetric.

\subsection{Spatial Toeplitz Inverse Covariance-Based Clustering}
\subsubsection{Representing Dependencies among Attributes with Markov Random Field}
Places with similar attributes should be grouped into the same cluster.
To represent dependencies among multiple attributes of geographic objects, a Markov random field (MRF) is constructed for each cluster, in which each node in the network corresponds to an attribute (variable), and each edge indicates the dependency between different attributes (variables). Each geographic object is represented as a layer in MRF. The advantages of using MRF are two-fold.

First, edges in an MRF can be loosely interpreted as the partial correlations of a given pair of variables conditioned on the remaining variables \citep{rue2005gaussian,koller09probabilistic}. 
That is, any pair of variables in the network is non-adjacent if their computed partial correlation is zero, which means that an edge exists between two variables only if they are conditionally dependent given the remaining variables. 
Therefore, MRF is a powerful tool for modeling dependencies between different variables, as partial correlations control for the effect of the potential confounding variables. This is unlike the standard correlation method that does not take confounders into account.

Second, when using the MRF network, how a variable may affect other variables, conditioning on the remaining variables, is illustrated through its adjacencies, which provides interpretable insights to demonstrate the characteristics of clusters.
For instance, assume that we want to identify urban functional zones in cities, the MRF structure of a cluster that is identified as ``commercial regions" in a city may illustrate the partial correlations between two variables, e.g., the number of shopping malls and the number of restaurants, in a subregion. 
Suppose in the MRF structure that we identify a high partial correlation between the number of shopping malls of the center geographic object and the number of restaurants of its nearest geographic object. 
This implies that after controlling all the other variables, such as number of residential buildings, number of schools, number of hospitals, etc., these two variables are highly correlated. 
It is worth noting that there is a distinction between this notion of partial correlation and the standard correlation. 
The former one, as described, takes into account the confounding variables, and therefore may be less susceptible to spurious relationship.

As mentioned above, the \textit{subregions} of each geographic object are constructed for spatial clustering with consideration of spatial dependence, which serve as basic units for the MRFs. 
Hence, such MRF structures could have multiple layers (i.e., each layer represents a geographic object, and its multivariate values and their dependencies are represented as graph nodes and edges respectively) defined by the number of objects $R$ in subregions. The edges of graphs are both within a layer and across different layers, which correspond to the intradependencies and interdependencies of the object attributes, respectively.
An example is shown in Figure \ref{fig:idea}, in which the cluster's MRF contains three layers that represents the geographic object itself, the nearest object, and the second nearest object, respectively.
For each cluster identified, the partial correlation structure of all geographic objects inside subregions of this cluster is depicted. 
It should be noted that each subregion is grouped purely based on the attribute dependency structure of the cluster to which the center geographic object belongs.
Hence, the clusters are spatial-invariant.
That is to say, when assigning each geographic object to the cluster, the starting position of each subregion does not matter.

For each cluster, since the dependencies between attributes (variables) are different from other clusters, the MRF network structure is also different. For the $k$th cluster, we define its structure of MRF using the inverse covariance matrix of a multivariate Gaussian distribution, denoted as $\Theta_k\in \mathbb{R}^{DR\times DR}$. As described, the inverse covariance matrix illustrates the conditional independency structures among attributes inside a subregion.
By definition, if the entry $(\Theta_k)_{i,j}=0$, then these two attributes $i$ and $j$ are conditionally independent given the values of all other attributes in the subregion. 

Next, we will introduce the specific structure of the inverse covariance matrix that helps preserve spatial invariance, including the \textit{spatial Toeplitz matrix} in section \ref{sec:toeplitz_matrix} and a \textit{Toeplitz graphical lasso} strategy to estimate the inverse covariance matrix in section \ref{sec:lasso}.


\subsubsection{Spatial Toeplitz Matrix}\label{sec:toeplitz_matrix}
To help preserve spatial invariance within a subregion, we restrict the inverse covariance matrices to follow the block Toeplitz form (i.e., a special diagonal-constant matrix)~\citep{akaike1973block}. 
In particular, the $DR \times DR$ inverse covariance matrix of each cluster is defined as
\begin{equation}
    \Theta_k=
    \begin{bmatrix}
A_k^{(0)} & (A_k^{(1)})^T & (A_k^{(2)})^T & \ldots & \ldots & (A_k^{(R-1)})^T\\
A_k^{(1)} & A_k^{(0)} & (A_k^{(1)})^T & \ddots & & \vdots\\
A_k^{(2)} & A_k^{(1)} & A_k^{(0)} & \ddots& \ddots&\vdots\\
\vdots  &  \ddots& \ddots &\ddots & (A_k^{(1)})^T & (A_k^{(2)})^T \\
\vdots  &       & \ddots & A_k^{(1)} & A_k^{(0)} & (A_k^{(1)})^T \\
A_k^{(R-1)} & \ldots & \ldots & A_k^{(2)}  & A_k^{(1)}  & A_k^{(0)} \\
\end{bmatrix},
\end{equation}
where $A_k^{(0)}, A_k^{(1)}, \dots , A_k^{(R - 1)} \in \mathbb{R}^{D\times D}$.
Both \textit{intradependencies} and \textit{interdependencies} are indicated by the Toeplitz matrix.
The former refers to the sub-block $A^{(0)}_k$ whose entry $(A^{(0)}_k)_{i,j}$ represents the partial correlation between attributes $i$ and $j$ in the same geographic object within the $k$th cluster.
Additionally, the off-diagonal sub-blocks refer to the relationships between the attributes of different geographic objects in the subregion.
For example, suppose a geographic object $x_n$ belong to the $k$th cluster. Then, $(A^{(1)}_k)_{i,j}$ indicates the relationship between the attribute $i$ of $x_n$ and the attribute $j$ of its nearest neighbor $x_n^{(1)}$;
similarly, $A^{(r)}, 1 < r < R$ shows the edge structure between the geographic object $x_n$ and its $r$th nearest neighbor $x_n^{(r)}$.
Using the block Toeplitz structure of the inverse covariance, we could characterize the relationships between multiple attributes across space with the spatial-invariance assumption that each geographic object only depends on its $(R-1)$ nearest neighbors regardless of its absolute location. Note that such attribute dependencies may vary across space but we assume it is relatively stable within each cluster.

\subsubsection{Overall Optimization Problem}\label{sec:overall_optimization}
After introducing the MRF and the Toeplitz matrix, the overall spatial clustering problem can be defined as an optimization problem. 
The objective is to solve for these $K$ inverse covariances $\boldsymbol\Theta=\{\Theta_1, \dots, \Theta_K\}$ for all clusters, and the assignment sets for the geographic objects $\boldsymbol P=\{P_1, \dots, P_K\}$ with $P_i \subset  \{1, 2, \dots, N\}$.
It is termed as the \textit{spatial Toeplitz inverse covariance-based clustering} (STICC), which solves the following optimization problem:
\begin{equation}\label{eq:sticc_problem}
   \min_{\boldsymbol\Theta \in \mathcal{T},\boldsymbol P} \sum_{k=1}^K
   \left[
   \displaystyle\sum_{X_n \in P_k}
   \left(
   \overbrace{-\mathcal{L}(\Theta_k;X_n)}^{\text{log likelihood}}+
   \overbrace{\beta \mathbbm{1} \{ X_n^{(1)}\notin P_k\} }^{\text{spatial consistency}}
   \right
   )
    +\overbrace{\|\lambda \odot\Theta_k \|_{1, \operatorname{off}}}^{\text{sparsity}}
   \right],
\end{equation}
where $\mathcal{T}$ denotes the set of $DR \times DR$ matrices that are symmetric block Toeplitz.
For each geographic object $x_n$, recall that $X_n$ is defined as the subregion in which $x_n$ is the center point.
In the optimziation problem, each subregion $X_n$ is assigned to one cluster, and $-\mathcal{L}(\Theta_k;X_n)$ denotes the negative log likelihood that the subregion $X_n$ belongs to the $k$th cluster.
The cluster assignment of the subregion $X_n$ is used as that of the geographic object $x_n$.

Here, $\mathbbm{1} \{ X_n^{(1)}\notin P_k\}$ is a spatial consistency indicator function that determines if $X_n$ and its nearest subregion $X_n^{(1)}$ are in the same group, defined as
\begin{equation}
    \mathbbm{1} \{ X_n^{(1)}\notin P_k\}=
    \begin{cases}
      0, & \text{if}\ X_n, X_n^{(1)} \in P_k, \\
      1, & \text{otherwise.}
    \end{cases}
\end{equation}
$\beta$ is a hyperparamter that controls the importance of this term.
This penalty term is used to encourage the neighboring subregions $X_n$ and $X_n^{(1)}$ to be assigned to the same cluster, so that the spatial contiguity is maintained.

An $\ell_1$ penalty term $\|\lambda \odot \Theta_k\|_{1,\operatorname{off}}$ is incorporated to enforce the sparsity on the off-diagonal entries of the inverse covariance matrices, where $\lambda \in \mathbb{R}^{DR \times DR}$ is a hyperparameter that controls the sparsity level, and $\odot$ denotes the entry-wise product.
This penalty term helps enforce sparsity on the edges of the corresponding MRFs, since the non-zero entries in the inverse covariance matrices correspond to the underlying structure of the MRFs.
Note that the sparsity assumption, or more specifically, the $\ell_1$ constraint, has been adopted and is essential in many areas such as compressed sensing \citep{donoho2006compressed,candes2005decoding}, linear regression \citep{tibshirani1996lasso,zou2006adaptive}, and  graphical model selection \citep{friedman2008sparse,aragam2015concave,ng2020role}. This is especially useful for high-dimensional tasks with a large number of variables and limited samples \citep{hastie2015statistical}, as is the case in our multivariate spatial clustering.

Since we focus on the Gaussian inverse covariance matrices and assume that the joint distribution of $X_n$ follows a multivariate Gaussian distribution, the log likelihood $\mathcal{L}(\Theta_k;X_n)$ is given by
\begin{equation}\label{eq:likelihood}
\mathcal{L}(\Theta_k;X_n) = -\frac{1}{2}(X_n-\mu_k)^T\Theta_i(X_n-\mu_k) +\frac{1}{2}\log\det\Theta_k-\frac{DR}{2}\log2\pi,
\end{equation}
where $\mu_k$ refers to the empirical mean of the $k$th cluster, and $\det\Theta_k$ denotes the determinant of the matrix $\Theta_k$.


\subsection{Cluster Assignments and Parameter Updates}
\label{sec:clusterassignments}
\begin{algorithm}
\caption{Overall steps for STICC}\label{alg:algorithm1}
\begin{algorithmic}
\State \textbf{initialize} cluster assignments $\boldsymbol P$ and cluster parameters $\boldsymbol\Theta$
\State \textbf{while} not stationarity 
\State E-step: cluster assignments $\rightarrow\boldsymbol P$ 
\State M-step: parameter updates $\rightarrow\boldsymbol\Theta$
\State \textbf{endwhile}
\State return $\boldsymbol P$, $\boldsymbol\Theta$
\end{algorithmic}
\end{algorithm}

To solve the problem defined in equation \eqref{eq:sticc_problem} that involves both combinatorial and continuous optimization, we adopt an approach similar to the expectation-maximization (EM) method that alternates between two steps: cluster assignments and cluster parameters updates.
As shown in Algorithm \ref{alg:algorithm1}, the overall procedure of the STICC algorithm is as follows:
\begin{enumerate}
\item \textbf{Initialization:} initialize cluster assignments $\boldsymbol P$ and cluster parameters $\boldsymbol\Theta$. The former could be randomly initialized, or initialized using other clustering methods, if available, such as K-Means or Gaussian mixture model.
\item \textbf{E-step:} compute the set of cluster assignments $\boldsymbol P$ for the geographic objects.
\item \textbf{M-step:} update cluster parameters $\boldsymbol\Theta$.
\item If the algorithm converges, then stop; otherwise, repeat steps (2) and (3).
\end{enumerate}
Further details of the E-step and M-step are described in the following sections.



\subsubsection{E-Step: Cluster Assignment with Spatial Consistency}
In this step, we aim to assign each geographic subregion, $X_1, \dots, X_N$, to one of the $K$ clusters given the fixed cluster parameters $\Theta_i, i= 1,2,\dots,K$ (i.e., inverse covariances).
The number of clusters $K$ should be assigned manually.
In practice, $K$ can be inferred by using methods for unsupervised clustering such as silhouette score \citep{ogbuabor2018clustering}, elbow method \citep{syakur2018integration}, information criterion approach \citep{kodinariya2013review}, and information-theoretic approach \citep{sugar2003finding}.
Particularly, the optimization problem for assignment of subregions to clusters is derived from the equation \eqref{eq:sticc_problem} and given by
\begin{equation}\label{eq:sticc_subproblem}
   \displaystyle\min_{\boldsymbol P}\sum_{k=1}^K\sum_{X_n \in P_k}
   \left(
   \overbrace{-\mathcal{L}(\Theta_k;X_n)}^{\text{log likelihood}}+
   \overbrace{\beta \mathbbm{1} \{ X_n^{(1)}\notin P_k\} }^{\text{spatial consistency}}
   \right
   ).
\end{equation}
As described in Section \ref{sec:overall_optimization}, this formulation aims to maximize both the log likelihood and the spatial consistency.

\begin{figure}[!t]
    \centering
    \includegraphics[width=0.8\textwidth]{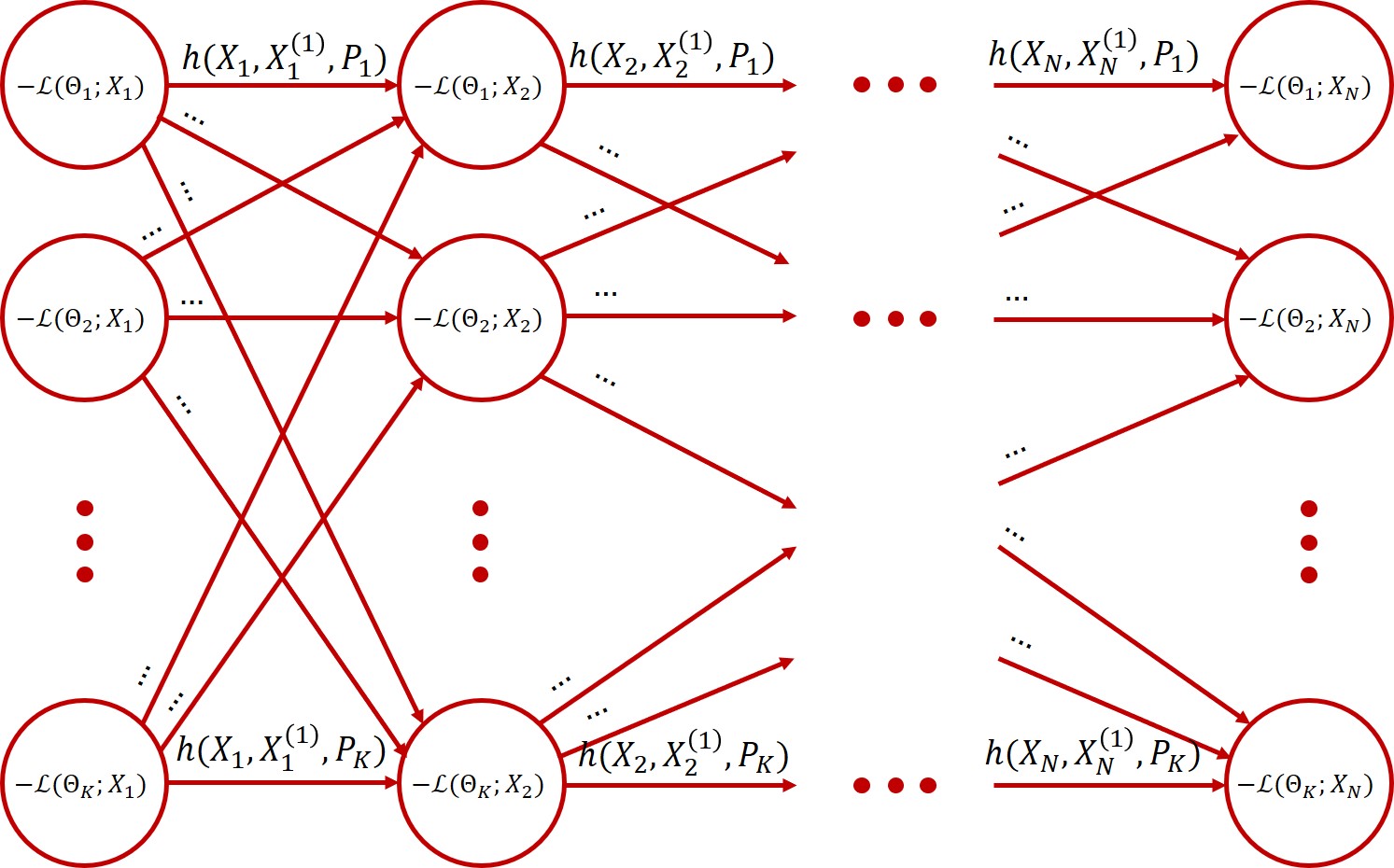}
    \caption{
    Problem \eqref{eq:sticc_subproblem} can be converted to a minimum cost path finding task from subregion $1$ to $N$, where the node cost is the negative log likelihood of that point being assigned to a given cluster, and the edge cost is determined by the function whether the $n$th geographic object and its nearest neighbor belong to the same cluster; if so, then it is $0$, otherwise, a penalty $\beta$ is added.}
    \label{fig:dp}
\end{figure}

A naive approach to solve the above combinatorial optimization problem is by enumerating all possible assignments of the subregions to the clusters, which, however, quickly becomes infeasible as there are $K^N$ possible combinations, leading to exponential increase of running time in the number of geographic objects.
To handle this issue, we leverage a dynamic programming approach to solve the problem with a running time $O(KN)$, as illustrated in the Figure \ref{fig:dp}.
Such a problem can be cast into a minimum cost path finding task \citep{viterbi1967error} from subregion $X_1$ to $X_N$.
As shown in Figure \ref{fig:dp}, the node cost refers to the negative log likelihood $-\mathcal{L}(\Theta_k;X_n)$ of a specific cluster that the geographic object is assigned to.
We define the edge cost from the node $-\mathcal{L}(\Theta_{k};X_n)$ to the destination node $-\mathcal{L}(\Theta_{k^{'}};X_{n+1})$ to be $h(X_{n}, X_{n}^{(1)},P_{k})$ that is decided based on the cluster assignment of the nearest subregion to $X_{n}$, and formally defined as
\begin{equation}
    h(X_{n}, X_{n}^{(1)},P_{k})=
    \begin{cases}
      0, & \text{if}\ X_{n}, X_{n}^{(1)} \in P_{k}, \\
      \beta, & \text{otherwise.}
    \end{cases}
\end{equation}
For example, in terms of the subregion $X_1$, if its nearest subregion $X_1^{(1)}$ belongs to the same cluster, then there is no additional cost except for the negative log likelihood; otherwise, the penalty $\beta$ should be added.
The pseudocode for cluster assignment is illustrated in Algorithm \ref{alg:cluster_assignment}.

\begin{algorithm}
\caption{E-step: Cluster Assignment}\label{alg:cluster_assignment}

\begin{algorithmic}
\State \textbf{Input}: $\{X_n\}_{n=1}^{N}$: a set of $N$ subregions; 
\State $\{X_n^{(1)}\}_{n=1}^{N}$: a set of $N$ subregions that indicates the nearest neighbor of $X_n$; 
\State $\mathcal{L}(\Theta_k;X_n)$: negative log likelihood of the subregion $X_n$ that belongs to cluster $k$, for $k=1,2,\dots,K$ and $n=1,2,\dots,N$; 
\State $\beta >0$: penalty coefficient.
\State \textbf{initialize} FutureCostVals $\coloneqq$ list of $N$ lists of $K$ zeros.
\For{$n=N-1,\dots,1$}
\State $m\coloneqq$ index of $X_n^{(1)}$
\State FutureCost $\coloneqq$ FutureCostVals[$m$, $:$]
\State LogLikelihoodVals $\coloneqq$ $\left(\mathcal{L}(\Theta_1;X_n^{(1)}),\mathcal{L}(\Theta_2;X_n^{(1)}),\dots,\mathcal{L}(\Theta_K;X_n^{(1)})\right)$
\For{$k=1,\dots,K$}
\State TotalVals $=$ FutureCost $+$ LogLikelihoodVals $+$ $\beta$
\State TotalVals[$k$] $=$ TotalVals[$k$] $-$ $\beta$
\State FutureCostVals[$n$, $k$] $=$ $\min$(TotalVals)
\EndFor
\EndFor
\State Path $\coloneqq$ list of $N$ zeros.
\For{$n=1,\dots,N$}
\State $m\coloneqq$ index of $X_n^{(1)}$
\State FutureCost $\coloneqq$ FutureCostVals[$m$, $:$]
\State LogLikelihoodVals $\coloneqq$ $\left(\mathcal{L}(\Theta_1;X_n^{(1)}),\mathcal{L}(\Theta_2;X_n^{(1)}),\dots,\mathcal{L}(\Theta_K;X_n^{(1)})\right)$
\State TotalVals $=$ FutureCost + LogLikelihoodVals $+$ $\beta$
\State TotalVals[Path[$n$]] = TotalVals[Path[$n$]] $-$ $\beta$
\EndFor
\State \textbf{return} Path
\end{algorithmic}
\end{algorithm}

As suggested by the objective function \eqref{eq:sticc_subproblem}, setting $\beta$ to $0$ reduces to independently assigning each subregion $X_1, \dots, X_N$ to a certain cluster based on its log likelihood, which, as explained in Section \ref{sec:intro}, is not desirable as it is not much different from attribute-based clustering that does not adequately incorporate the geographic information.
With the increase of $\beta$, nearby subregions are encouraged to be allocated to the same cluster to keep the spatial contiguity.
For the extreme case where $\beta \rightarrow \infty$, all subregions will be grouped into the same cluster as the penalty term dominates entirely.



\subsubsection{M-Step: Cluster Parameter Updates with Toeplitz Graphical Lasso}
\label{sec:lasso}
The other step is the parameter updates given the fixed cluster assignments $\boldsymbol P$.
It aims to update parameters $\Theta_k$ of cluster $k= 1,\dots,K$.
Here, we assume that the multivariate Gaussian distribution of $X_n$ has zero-mean, and rewrite the log likelihood term in Eq. \eqref{eq:likelihood} as
\begin{equation}
   \displaystyle\sum_{X_n \in P_k}
   \mathcal{L}(\Theta_k;X_n)
   =-|P_k|( \log \det \Theta_k + \operatorname{tr}(S_k \Theta_k)) + C,
\end{equation}
where $|P_k|$ indicates the number of geographic objects assigned to cluster $k$, $S_k$ denotes the empirical covariance of these objects, $\operatorname{tr}$ refers to tracing over the diagonal elements of the matrix, and $C$ is a constant.

To update the parameter $\Theta_k$, the optimization subproblem of the M-step can be  written as
\begin{equation}
\min_{\Theta_k \in\mathcal{T}} \quad -\log\det \Theta_k + \operatorname{tr}(S_k\Theta_k) + \frac{1}{|P_k|} \|\lambda \odot \Theta_k\|_{1,\operatorname{off}}.
\end{equation}
The above problem, termed as the \textit{Toeplitz graphical lasso} \citep{hallac2017toeplitz}, is convex, as is the case in the typical setting of graphical lasso \citep{friedman2008sparse}.

A straightforward approach to solving this optimization problem is to use the standard coordinate descent method developed by \citet{friedman2008sparse}, which, however, may not scale up to large inverse covariances, especially since our EM-like procedure requires solving this problem up to tens or hundreds of times before converging to a stationary solution.
Other improved alternatives include the second-order approach with quadratic approximations \citep{hsieh2014quic}, or the alternating direction method of multipliers (ADMM) \citep{boyd2011admm,scheinberg2010sparse}.
In this work, we adopt the ADMM method, a distributed algorithm that solves an optimization problem by decomposing it into smaller subproblems, each of which are easier to solve.
Such a strategy is also used in the TICC algorithm \citep{hallac2017toeplitz} and more technical details can be found there.

\section{Experiment}
We have implemented the proposed STICC algorithm in Python.
The PySal library is used for constructing the spatial relationships, performing the k-nearest neighbor spatial matrix, and constructing the Delaunay triangulation \citep{rey2010pysal}.
Following the reproducibility and replicability guidelines \citep{wilson2020five}, the code for this project is open-sourced and available on a public repository on GitHub: \url{https://github.com/GeoDS/STICC}.
In the following sections, the developed STICC algorithm is applied in two case studies, a synthetic experiment and a real-world scenario to demonstrate how this method can be used in spatial clustering to provide meaningful insights for repeated geographic pattern discovery.

\subsection{Experiment Set Up}
In this section, we discuss the common settings of the two experiments, including baseline methods and evaluation metrics.
Both experiments are performed on a cloud server with Ubuntu 16.04 system and Python 3.8 version.

\subsubsection{STICC and Baseline Methods}
There are two groups of clustering methods used in our experiments.
The first group includes a series of the proposed STICC clustering algorithms with different parameter settings on $R$ and $\beta$, so that we can explore the performance of the algorithm under different conditions.
We start with a fixed $\beta$ and an increasing $R$ value from $R=1$ to $R=4$.
When $R=1$, no nearby geographic objects are used for the construction of subregions.
Then, by picking up the best $R$, we test different $\beta=0,1,3,5$, respectively.
It should be noted that when $\beta=0$, there is no spatial consistency strategy performed.

The second group contains the following six commonly used unsupervised clustering algorithms to serve as baseline approaches.

\begin{enumerate}
    \item K-Means: The K-Means clustering algorithm with Euclidean distance in a multivariate space (including only aspatial attributes in our experiments) \citep{anderberg1973cluster}.
    \item Spatial K-Means: It is an adaption of the K-Means that treats the coordinates of each object as two attributes together with other aspatial attributes.
    \item GMM: The Gaussian mixture model aims at representing the probability distribution of each object for clustering \citep{rasmussen1999infinite}.
    \item CURE: The clustering using representatives model that extends K-Means clustering approach and is more robust to outliers, shape and size variances \citep{guha1998cure}.
    \item DBSCAN: The density-based spatial clustering of applications with noise \citep{ester1996density}.
    \item Spatially Constrained Multivariate Clustering: 
    The spatially constrained multivariate clustering algorithm that uses minimum spanning tree and a strategy termed as SKATER that can identify natural clusters and evidence accumulation to evaluate the probability of cluster membership \citep{assunccao2006efficient}.
\end{enumerate}
The first four methods are attribute-based clustering algorithms; DBSCAN algorithm is a density-based clustering approach; and the spatially constrained multivariate clustering approach is a regionalization-based approach.
Therefore, the characteristics of the three-category spatial clustering approaches can be depicted comprehensively and compared with our proposed STICC method. 
All algorithms except GMM and spatially constrained multivariate clustering are performed using a python package \textit{pyclustering} \citep{novikov2019pyclustering}; the GMM is performed using the package \textit{sklearn}; the spatially constrained multivariate clustering is executed in ArcGIS Pro 2.8.
We use default settings of each algorithm as they mainly require the number of clusters $K$ only.
A grid-search strategy is performed for DBSCAN to infer optimal results, since this algorithm requires two input parameters, namely the search \textit{radius} and the minimum number of data points \textit{minPts} in each cluster.


\subsubsection{Evaluation Metrics}
To evaluate the performances of all these above-mentioned clustering methods, we use the following three metrics from different aspects: 
adjusted rand index (ARI), macro-F1, and join count ratio.
Generally, clustering is treated as an unsupervised learning process.
However, since we also know the ``ground-truth'' results of the clustering in our case studies, it can be seen as a supervised multi-class classification problem as well.
The ARI and macro-F1 are good indicators in measuring the accuracy of clustering results.
While the join count ratio can be used to measure the spatial dependence.
Given that the STICC method and several baseline methods may acquire the appropriate number of clusters \textit{K}, we assign the ``true" number of clusters of each dataset as \textit{K}.

The ARI measures the similarity between two clustering results by counting pairs of samples that are assigned with the same or different clusters \citep{steinley2004properties}.
A value between 0 and 1 is produced for each round of comparison between clustering results.
The higher the value, the more similar the two clustering results.

The macro-F1 score has been widely used in the field of machine learning for measuring algorithm performance \citep{fujino2008multi}.
The precision and recall of each cluster are computed at first to infer the F1 scores.
The average of all F1 scores for all clusters are calculated as the macro-F1 score for making comparisons between the STICC method with other baseline models.
Such an index achieves the balance between precision and recall.
The macro-F1 score ranges from 0 to 1 as well. 
A value near 1 indicates a better clustering result.
To obtain the macro-F1 score, given that we do not know the one-to-one cluster label matching between the ``ground truth'' and the output clustering results, all potential permutations are enumerated and the macro-F1 score of each item is computed.
The permutation with the highest macro-F1 score is used as the final clustering labels.

Researchers have made extensive use of the former two indices for measuring the performance of clustering approaches.
Recall that an ideal spatial clustering result may achieve a balance between attribute similarity and spatial contiguity. Hence, a quantitative examination of the degree of spatial contiguity that is preserved in clustering results is necessary. In other words, how many nearby geographic objects tend to belong to the same cluster. Inspired by the join count statistics \citep{dacey1965review,cliff1973spatial}, we measure the proportion of connections where neighboring geographic objects belong to the same cluster to all connections among adjacent geographic objects, which is termed as textit{join count ratio}.
For a pair of two neighboring geographic objects, their resulting cluster labels might be the same or different. It should be noted that, for polygons, neighboring geographic objects refer to adjacent polygons or lattices with shared boundaries; while for points, Delaunay triangulation or other types of connectivity among points should be constructed first, and neighboring points are considered as adjacent geographic objects. We adopt the following equations to calculate the join count ratio:
\begin{equation}
    J = J_\text{same} + J_\text{diff}
\end{equation}
\begin{equation}
    J_\text{ratio} = \frac{J_\text{same}}{J}
\end{equation}
where $J_\text{same}$ is the value of join counts that neighboring objects belong to the same cluster, $J_\text{diff}$ indicates the value of join counts that neighboring objects belong to different clusters.
$J_\text{ratio}$ ranges from 0 to 1, and the higher the proportion of neighboring geographic objects belonging to the same category, the higher spatial dependence of output results.

Such a measure may help us examine the spatial contiguity within each subregion given the spatial dependency of attributes.
We acknowledge that there are multiple other measures used for examining spatial contiguity of geographic patterns, such as Moran's I and Geary's C.
However, they are inappropriate for nominal data (e.g., clustering labels) in this work.

\subsection{Synthetic Experiment}
\subsubsection{Dataset Preparation}
\begin{figure}[h]
    \centering
    \includegraphics[width=0.7\textwidth]{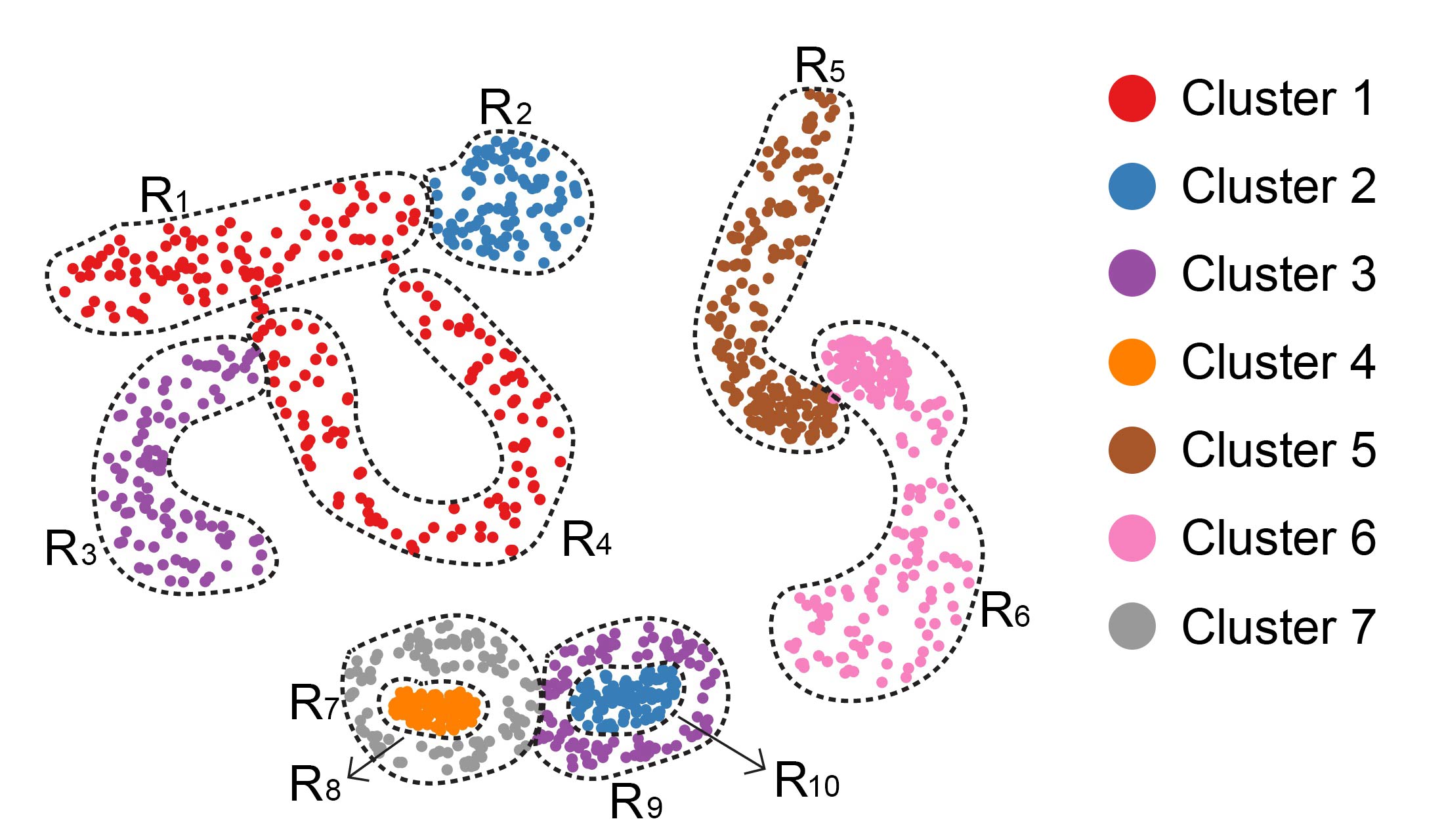}
    \caption{The generated synthetic dataset with ten regions ($R_1, R_2, \dots, R_{10}$) that belong to seven clusters.}
    \label{fig:exp1_gt}
\end{figure}

We first generate a synthetic point dataset with multiple attributes to test the performance of our proposed STICC method. 
Inspired by previous works \citep{estivill2002argument,nosovskiy2008automatic,liu2012density,mai2018adcn}, the spatial distribution of points are generated as shown in the Figure \ref{fig:exp1_gt} with the following characteristics.
According to the figure, regions $R_1$ and $R_4$, regions $R_2$ and $R_{10}$, and regions $R_3$ and $R_9$, belong to the same clusters, respectively, but they may locate in different positions and are not spatially adjacent to each other (e.g., $R_3$ and $R_9$, $R_2$ and $R_{10}$).
These regions have diverse densities of points.
For example, regions 4 and 8 have uniform point densities, while point densities fluctuate in regions 5 and 6; densities of points in adjacent regions may be similar (e.g., regions 7 and 9) or different (such as regions 7 and 8).
Also, the minimum distances between two nearby regions can be different.
For instance, region 5 is closely connected (i.e., ``touch") with region 6 while is not directly adjacent to region 2.

Though there are ten regions of points in total, they only belong to seven predefined clusters according to their multiple attribute similarity.
For each point, we generate five attributes as its multi-dimensional features.
The attribute values of each point in a specific cluster are randomly assigned in a certain range, and such a range follows normal distributions with distinct settings of mean $\mu$ and standard error $\theta$.
After randomly generating these parameters for multiple times, similar conclusions can be obtained from our further experiments. 
Hence, we only show one group of parameters as an example in Table \ref{tab:exp1_gt} for demonstrating the experiment settings.
Table \ref{tab:exp1_gt} provides the means $\mu$ and standard errors $\theta$ of the normal distribution for the attribute value range of objects in each cluster.
Attribute values may overlap in different clusters due to the settings of $\theta$.
Given that different attributes of one geographic object may have diverse value ranges in the real world (e.g., the average temperature and the population of a place are unrelated), we assign disparate settings of normal distributions to create these attributes.
As suggested in Table \ref{tab:exp1_gt}, the attribute A ranges from 1-7 while the attribute D varies from 400-1000.

\begin{figure}[H]
    \centering
    \includegraphics[width=0.9\textwidth]{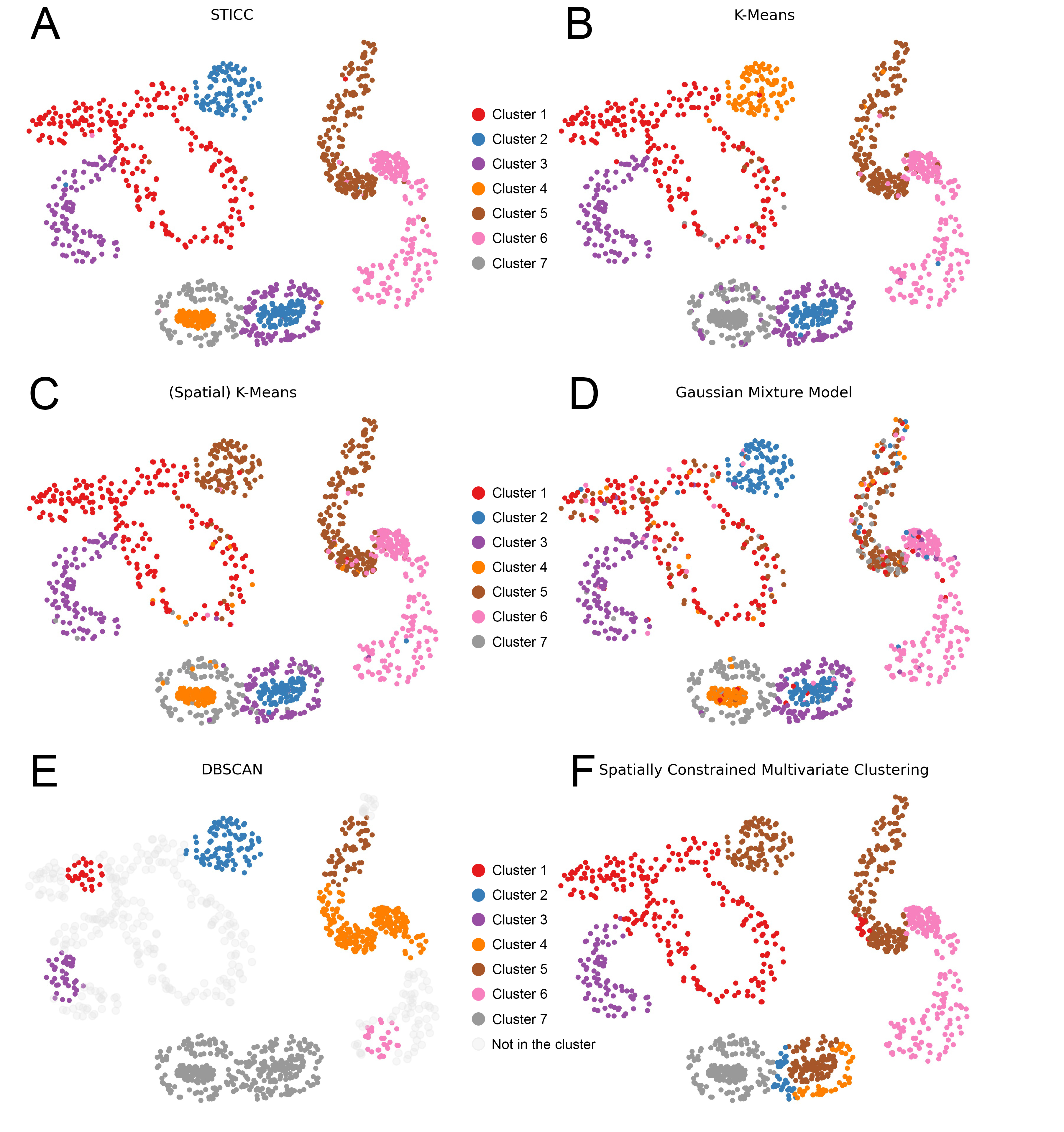}
    \caption{Results of the synthetic dataset (seven clusters). (A) the proposed STICC method (with $r=3$ and $\beta=3$; (B) K-Means; (C) spatial K-Means; (D) Gaussian mixture model; (E) DBSCAN (with $radius=1250$ and $minPts=25$; (F) spatially constrained multivariate clustering.}
    \label{fig:exp1_result}
\end{figure}

\begin{table}[]
\caption{Settings of the generated synthetic dataset with five attributes.}
\resizebox{0.8\textwidth}{!}{%
\begin{tabular}{@{}lllllllllll@{}}
\toprule
\multirow{2}{*}{Cluster} & \multicolumn{2}{l}{Attribute A} & \multicolumn{2}{l}{Attribute B} & \multicolumn{2}{l}{Attribute C} & \multicolumn{2}{l}{Attribute D} & \multicolumn{2}{l}{Attribute E} \\ 
                        & $\mu$     & $\theta$                  & $\mu$     & $\theta$                   & $\mu$     & $\theta$                   & $\mu$     & $\theta$                   & $\mu$      & $\theta$                  \\ \cmidrule(r){1-11}
1                       & 4      & 1     & 1      & 3     & 80     & 20    & 1000   & 350   & 999     & 3    \\
2                       & 5      &  1                      & 7      &  3                      & 30     &  20                      & 900    &   350                     & 992     & 3                      \\
3                       & 6      &  1                      & 2      &  3                      & 20     &   20                     & 600    &   350                     & 1005    & 3                      \\
4                       & 1      &  1                      & 3      &  3                      & 100    &    20                    & 700    &   350                     & 1003    & 3                      \\
5                       & 3      &  1                      & 6      &  3                      & 60     &     20                   & 800    &   350                     & 999     & 3                      \\
6                       & 7      &  1                      & 4      &  3                      & 70     &      20                  & 400    &   350                     & 998     & 3                      \\
7                       & 2      &  1                      & 5      &  3                      & 40     &       20                 & 500    &   350                     & 1008    & 3                      \\ \bottomrule
\end{tabular}
}
\label{tab:exp1_gt}
\end{table}

\begin{table}[]
\caption{Results of different clustering approaches on synthetic dataset (number of clusters, $K$=7).}
\resizebox{\textwidth}{!}{%
\begin{tabular}{@{}llllll@{}}
\toprule
          & \multicolumn{2}{l}{Cluster}        & \multirow{2}{*}{Adjusted Rand Index} & \multirow{2}{*}{Macro-F1} & \multirow{2}{*}{Join Count Ratio} \\ \cmidrule(r){1-3}
          & R               & $\beta$             &                             &                             &                            \\ \cmidrule(l){4-6} 
STICC     & 1               & 3                & 0.894                       & 0.954                       & 0.851                      \\
          & 2               & 3                & 0.931                       & 0.973                       & 0.878                      \\
          & 3               & 3                & \textbf{0.960}                        & \textbf{0.984}                       & 0.901                      \\
          & 4               & 3                & 0.574                       & 0.550                        & 0.822                      \\
          & 3               & 0                & 0.947                       & 0.977                       & 0.888                        \\
          & 3               & 1                & 0.952                       & 0.981                       & 0.896                      \\
          & 3               & 5                & 0.818                       & 0.771                       & 0.886                      \\
Baseline Clustering Methods & \multicolumn{2}{l}{K-Means}        & 0.799                       & 0.735                       & 0.544                        \\
          & \multicolumn{2}{l}{CURE}           & 0.0006                      & 0.053                       & -                       \\
          & \multicolumn{2}{l}{Spatial K-Means} & 0.830                       & 0.744                       & 0.881                      \\
          & \multicolumn{2}{l}{GMM}            & 0.703                       & 0.850                       & 0.685                      \\
          & \multicolumn{2}{l}{DBSCAN (\textit{radius}=1250, \textit{minPts}=25)}         & 0.327                       & -                           & 0.933                      \\
          & \multicolumn{2}{l}{Spatially constrained multivariate clustering}         & 0.629                       & 0.546                           & \textbf{0.936}                      \\\bottomrule
\end{tabular}
}
\label{tab:exp1_result}
\end{table}

\subsubsection{Clustering Results}
The comparison results of different clustering methods are shown in Table \ref{tab:exp1_result}.
Several selected clustering results are displayed in Figure \ref{fig:exp1_result}.
To guarantee that model performances are compared under same conditions, we generate identical synthetic data for each of the methods.

\textbf{Accuracy}: 
In accordance with the Table \ref{tab:exp1_result}, our proposed STICC method significantly outperforms all baseline clustering approaches in identifying such dispersed distributed similar subregions regarding both ARIs and macro-F1 scores. The STICC method with different parameters achieves very high accuracy in most cases.
The ARIs of all experiments but one are higher than 0.81, and the macro-F1 scores in five out of seven experiments are over 0.95 with STICC.
The model with parameters $R=3$ and $\beta=3$ performs the best.
Its macro-F1 score reaches 0.984.
Figure \ref{fig:exp1_result}(A) shows the results of STICC ($R=3, \beta=3$). 
As shown in the figure, all clusters are correctly identified and the majority of points are classified to the correct group.
Repeated clusters (e.g., $R_3$ and $R_9$, $R_2$ and $R_{10}$), even if located in different positions, are detected accurately.
Only a small proportion of points are labeled incorrectly (as indicated by the macro-F1 score, more than 98\% points are classified correctly).

As for the baseline approaches, when setting the number of clusters $K$=7,  spatial K-Means has the best ARI of 0.83 (0.13 less than the best STICC), and GMM has the highest macro-F1 of 0.850 (0.134 less than the best STICC).
Figures \ref{fig:exp1_result}(B)-(D) depict the results of those baseline clustering approaches.
Results of K-Means show that most clusters are identified (Figure \ref{fig:exp1_result}(B)).
However, regions 7 and 8 are grouped into the same cluster incorrectly; regions 2 and 10, which belong to the same group, are misclassified.
According to Figure \ref{fig:exp1_result}(C), coordinates of points are treated as two attributes when executing the spatial K-Means clustering, nearby points are encouraged to be grouped into the same cluster.
In the synthetic dataset, such a method performs worse than K-Means as regions 2 and 5 are misclassified into the same group.
GMM, as shown in Figure \ref{fig:exp1_result}(D), generally detects all major clusters by attributes correctly. 
However, in comparison with the STICC method, the clustered points are in a more disordered way without good spatial contiguity.
In terms of density-based clustering algorithm (DBSCAN), according to Figure \ref{fig:exp1_result}(E), it can only discover densely-located point clusters based on locations but not attributes. 
Therefore, many points are not assigned into any clusters but as noise points, resulting in a relatively low ARI and a failure in getting a macro-F1 score as more than 7 clusters are obtained.
For the spatially constrained multivariate clustering using ArcGIS Pro, such a method can detect certain clusters (Figure \ref{fig:exp1_result}(F).
When performing the clustering algorithms, the Delaunay triangulations are constructed among points.
Only topologically nearby points are assigned into the same group and no repeated patterns are discovered.

\textbf{Spatial contiguity}:
According to Table \ref{tab:exp1_result}, the spatial contiguity (measured by the join count ratio) is well maintained by our proposed STICC method.
When $R=2, \beta=3$, and $R=3, \beta=0,1,3,5$, the join count ratio of all these five experiments are higher than that of the K-Means and GMM approaches.
It should be noted that a higher join count ratio is not necessarily to a higher ARI and macro-F1, as they measure different aspects of the model performance.
For example, if we only compare the K-Means algorithm and the spatially constrained multivariate clustering approach, the latter one has a higher join count ratio value but with lower ARI and macro-F1 score.
It makes sense as illustrated in Figure \ref{fig:exp1_result}(B) and \ref{fig:exp1_result}(C).
K-Means can identify correct clusters to a certain degree, though there are some noise points that are misclassified; while spatial K-Means incorrectly arrange the $R_2$ and $R_5$ together, as coordinates of points in these two regions are relatively similar.
In addition, both DBSCAN and the spatially constrained multivariate clustering approach have relatively high join count ratio, which indicates that the spatial contiguity is well preserved.
However, the ARI and macro-F1 scores, which measure classification accuracy, are relatively low for these two methods. This suggests that the join count ratio may only serve as a complement evaluation metric of ARI and marco-F1 scores from different perspectives.

\textbf{Parameters of STICC}:
We also try multiple combinations of $R, \beta$ values to see how these parameters affect clustering results based on our synthetic dataset.
With the increase of $R$ and fixing the $\beta$, in which more nearby points are used for constructing the subregions, the accuracy of clustering approaches improves at first, and then drops when $R=4$.
When we fix the $R$ nearest neighbors and change $\beta$, we can observe similar patterns (Table \ref{tab:exp1_result}).
The three indices increase at first and then drop when $\beta=5$.
A larger $\beta$ forces more nearby objects to be assigned into the same group.
 
In summary, the STICC approaches perform better than the other three groups of methods significantly in discovering repeated patterns of points with spatial contiguity maintained, and achieve the balance between spatial and aspatial attributes of geographic objects with appropriate parameters.

\subsection{Check-In Point Classification}
We then evaluate the STICC method as well as other baseline approaches on another real-world application: social media check-in point classification. 
The goal of this task is to extract semantic information such as place types from geographic coordinates and time stamps only.
It is very useful for multiple applications such as trajectory privacy protection \citep{rao2020lstm}, mobility pattern discovery \citep{soliman2017social}, and community detection \citep{zhao2016understanding}.

\subsubsection{Dataset Preparation}
Social media check-in datasets have been widely used for understanding and analyzing users' spatio-temporal activity patterns in location-based services (LBS). 
Users visit a place (usually referred to as a point of interest, POI), check in at that location, and post geotagged content.
Our testing data are from \citet{yang2014modeling}, which contain a Foursquare check-in dataset in New York City processed based on global social media check-in records. In this dataset, 118,316 social media check-in records from April 2012 to September 2013 were collected from Foursquare, containing 3,628 unique POIs and 500 users.
For each check-in record, the time stamp (including time of day and day of week), GPS coordinates of the POI, and its semantics represented by the POI categories are attached.
We use the POI category provided in the original dataset as the ground truth in this application.
This task aims at identifying POI categories based on temporal information of check-in records.
Given that some POIs have very limited visits, only those with at least 10 check-in points are kept.
We then extract days of the week $w \in (1,7)$, and the hour $h \in (0,23)$ information from time stamps as two attributes of each point. 
Given that check-in points attached to one POI have the same GPS coordinates, we add a random noise (following a uniform distribution) of every point so that each of them is shifted to locate at a different position while still inside the 100-meter radius of the center POI.
By doing so, the task becomes more challenging for clustering algorithms to correctly differentiate check-in points in mixed groups of sampling points.
After that, we test the STICC method as well as other baseline approaches to classify the category of check-in points on two tasks, home/work, and home/work/gym identification.
Again, the process is still an unsupervised learning problem without labels and the identification of place types requires temporal information (e.g., daytime vs. night time) after clustering. Existing studies have shown that the day of the week and the hour play important roles in identifying various place types \citep{wu2014intra,rao2020lstm}.
With only two attributes taken into account, we explore to what degree correct types of places can be inferred. 
Only attribute-based clustering methods are used for comparisons with STICC.
Density-based clustering methods can detect densely located point clusters.
However, it does not identify the semantics of each cluster.
Regionalization-based clustering methods may only partition the study area into two spatially adjacent groups.
Hence, the latter two methods are not appropriate for this application.

\subsubsection{Clustering Accuracy for Home/Work Identification}
\begin{figure}[h]
    \centering
    \includegraphics[width=\textwidth]{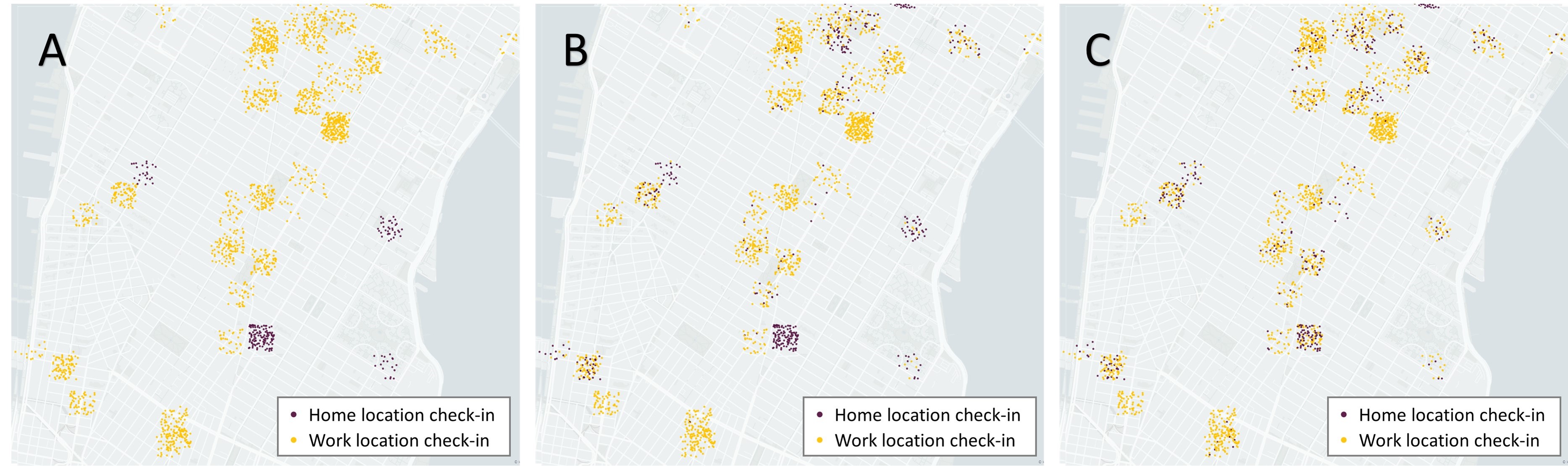}
    \caption{Results for home/work classification. (a) ground truth; (b) the proposed STICC method; (c) K-Means.}
    \label{fig:exp2_result}
\end{figure}

Home and work location detection is one of the fundamental tasks for LBS in practice.
Here, we use different clustering algorithms and perform a binary ``classification'' ($K=2$) to infer users' home and work locations from their check-in data.
The distributions of users' check-in points are plotted in Figure \ref{fig:exp2_result}(A) with yellow dots indicate home locations and purple points refer to work locations.
The clustering results of home/work identification of several clustering methods are demonstrated in Table \ref{tab:exp2_result}.
It can be inferred that the proposed STICC algorithm performs better than other baseline approaches using the three quality measurements.

\textbf{Accuracy:}
According to Table \ref{tab:exp2_result}, the ARI values of the STICC with different parameter settings are all over 0.35, and the macro-F1 scores are all higher than 0.79.
When $R=4,\beta=5$, the STICC has the highest ARI, macro-F1 score, and join count ratio.
In comparison, the K-Means is the best model with the highest ARI of 0.085, and GMM has the highest macro-F1 score of 0.587, both are far below the results from the STICC algorithm.
Three example maps are illustrated in Figure \ref{fig:exp2_result} to show the spatial distributions of check-in ground truth, STICC, and K-Means clustering results, respectively.
According to these maps, our developed STICC algorithm identifies home/work locations of most check-in points more correctly than the K-means, although there exist some mixed home/work clusters in the northern part on the map.

\textbf{Spatial contiguity:}
The join count ratio also illustrates that the STICC method maintains the spatial contiguity relatively well.
According to Table \ref{tab:exp2_result}, join count ratios are all greater than 0.80.
Also, as shown in Figure \ref{fig:exp2_result}(B), nearby check-in points are mostly grouped into the same cluster, which also suggests that spatial contiguity is well-preserved.
In comparison, attribute-based algorithms cannot keep the spatial contiguity well with relatively low join count ratio values (lower than 0.70).

\textbf{Parameters of STICC}:
Since the model performs best when $R=4$, multiple $\beta$ values are input into the STICC method to examine the resulting clustering maps for the POI classification task.
As shown in Table \ref{tab:exp2_result}, as the number of objects in each subregion increases, the accuracy and join count ratio of the STICC method also show increasing trends. 
Also, if $\beta$ increases, i.e., more nearby geographic objects are pushed to be in the same group, all three quality measurements increase as well.

\begin{table}[]
\caption{Results of different approaches for home and work classification (number of clusters, $K=2$).}
\resizebox{0.8\textwidth}{!}{%
\begin{tabular}{@{}llllll@{}}
\toprule
          & \multicolumn{2}{l}{Cluster}        & \multirow{2}{*}{Adjusted Rand Index} & \multirow{2}{*}{Macro-F1} & \multirow{2}{*}{Join Count Ratio} \\ \cmidrule(r){1-3}
          & R               & $\beta$             &                             &                             &                            \\ \cmidrule(l){4-6} 
STICC     & 1               & 3                & 0.390                       & 0.806                       & 0.829                      \\
          & 2               & 3                & 0.355                       & 0.792                       & 0.804                      \\
          & 3               & 3                & 0.433                       & 0.823                       & 0.834                       \\
          & 4               & 3                & 0.495                       & 0.844                       & 0.860                      \\
          & 4               & 0                & 0.445                       & 0.823                       & 0.822                      \\
          & 4               & 1                & 0.464                       & 0.834                       & 0.841                      \\
          & 4               & 5                & \textbf{0.514}                       & \textbf{0.850}                       & \textbf{0.871}                      \\
Traditional Clustering & \multicolumn{2}{l}{K-Means}        & 0.085                        & 0.321                       & 0.493                      \\
          & \multicolumn{2}{l}{CURE}           & 0.015                       & 0.578                        & 0.700                      \\
          & \multicolumn{2}{l}{Spatial-Kmeans} & 0.080                       & 0.384                        & 0.492                      \\
          & \multicolumn{2}{l}{GMM}            & 0.023                       & 0.587                       & 0.690                      \\ \bottomrule
\end{tabular}
}
\label{tab:exp2_result}
\end{table}

\subsubsection{Clustering Accuracy for Home/Work/Gym Identification}
We then conduct a follow-up experiment to perform multi-class ``classification'' ($K=3$) for grouping and identifying home, work, and gym locations. 
Since only the temporal patterns (hour and days of the week information) of check-in points are considered, it is an obstacle for most clustering algorithms even just extending the experiment from binary classification to three-class classification.
Table \ref{tab:exp2b_result} shows the three quality measurements of such a task. As mentioned above, all potential cluster label matching permutations are enumerated and the best results are reported here. 
Though the performances of all methods drop substantially, our proposed STICC algorithm continues yielding best results.

Generally speaking, similar conclusions can be obtained in comparison with the home/work classification task. 
As for ARI, the proposed STICC yields an ARI that is at least 0.12 higher than any other methods.
For macro-F1 scores, the STICC achieves the highest when $R=4, \beta=5$ which is higher than GMM (0.416) that performs the best among all other clustering approaches.
The resulting join count ratios of most STICC algorithms are higher than 0.64 with the highest being 0.712, while K-Means has the highest join count ratio with 0.675 among traditional clustering methods.
It is worth noting that when $R=4, \beta=0$, the join count ratio is comparatively low due to the lack of a spatial consistency strategy that forces nearby points to be in the same group.

\begin{table}[]
\caption{Results of different approaches for home/work/gym classification (number of clusters, $K=3$).}
\resizebox{0.8\textwidth}{!}{%
\begin{tabular}{@{}llllll@{}}
\toprule
          & \multicolumn{2}{l}{Cluster}        & \multirow{2}{*}{Adjusted Rand Index} & \multirow{2}{*}{Macro-F1} & \multirow{2}{*}{Join Count Ratio} \\ \cmidrule(r){1-3}
          & R               & $\beta$             &                             &                             &                            \\ \cmidrule(l){4-6} 
STICC     & 1               & 3                & 0.289                       & 0.476                       & 0.700                      \\
          & 2               & 3                & 0.204                       & 0.482                       & 0.647                      \\
          & 3               & 3                & 0.269                       & 0.500                       & 0.672                       \\
          & 4               & 3                & 0.298                       & 0.508                       & 0.700                      \\
          & 4               & 0                & 0.251                       & 0.495                       & 0.641                      \\
          & 4               & 1                & 0.273                       & 0.502                       & 0.669                      \\
          & 4               & 5                & \textbf{0.335}                       & \textbf{0.510}                       & \textbf{0.712}                      \\
Traditional Clustering & \multicolumn{2}{l}{K-Means}        & 0.041          & 0.352                       & 0.675                      \\
          & \multicolumn{2}{l}{CURE}           & 0.077                       & 0.294                        & 0.597                      \\
          & \multicolumn{2}{l}{Spatial-Kmeans} & 0.080                       & 0.397                        & 0.670                      \\
          & \multicolumn{2}{l}{GMM}            & 0.065                       & 0.416                       & 0.603                      \\ \bottomrule
\end{tabular}
}
\label{tab:exp2b_result}
\end{table}

Overall, both home/work and home/work/gym identification practices illustrate that our proposed STICC algorithm shows promising results in POI category identification problems inferred solely from hourly and days of the week.
Our results also suggest that it is necessary to consider nearby geographic observations in improving the model performance.

\section{Discussions}
\subsection{Influences of parameters}
The developed STICC algorithm requires four input hyperparameters $K$, $R$, $\beta$, and $\lambda$, which are manually determined.
To get optimal results, fine tuning is required when executing the algorithm. 
In this section, we briefly discuss the influence of these different parameters on the algorithm performance and hope to provide basic guidance in selecting hyperparameters.

$K$ refers to the number of clusters into which the data points expect to be grouped.
If there are some labeled ground-truth data, such data can be used for helping determine the $K$ value, as illustrated in the experiments shown in Section 3.
However, clustering algorithms are usually employed in solving unsupervised classification problems where no ground-truth data are provided.
Researchers may either need to manually select the appropriate $K$ relying on their experiences or based on the characteristics of datasets, or need to use silhouette score \citep{ogbuabor2018clustering}, elbow method \citep{syakur2018integration}, information criterion \citep{kodinariya2013review}, or information-theoretic  \citep{sugar2003finding} approaches to help determine the appropriate $K$.

\begin{figure}[h]
    \centering
    \includegraphics[width=0.5\textwidth]{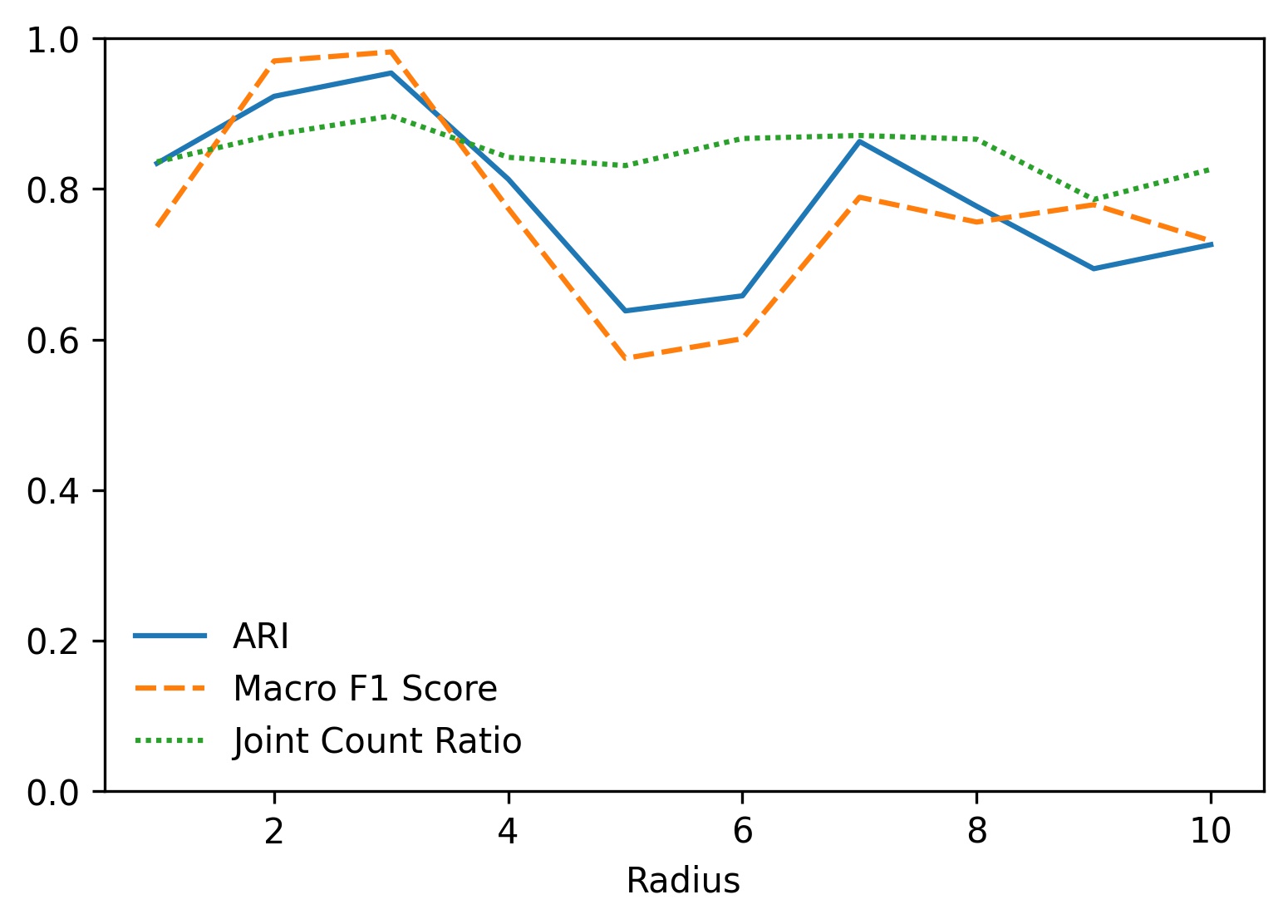}
    \caption{Changes of metrics for the synthetic dataset with different $R$.}
    \label{fig:radius}
\end{figure}

$R$ specifies the radius size of subregions, that is the number of geographic objects in each constructed subregion in our method. It also serves as the key parameter in integrating spatial context into the clustering algorithm.
The larger the $R$, the more neighbors are taken into account in clustering; and if $R=1$, only the center point itself is used.
The hypothesis for constructing subregions is that spatial dependencies among geographic objects within subregions are the same for a specific cluster and might be different across clusters.
Hence, choosing an appropriate $R$ value depends on the nature of the phenomenon being studied and the dataset.
For example, $R=$ 5 or 9 might be proper parameters for raster data (e.g., remote sensing images) considering the adjacency characteristics of image pixels (such as rook, bishop, and queen cases).
In general, a large $R$ might be suitable when objects are highly relevant to their neighbors.
With a large $R$, more geographic objects are taken into consideration when the STICC algorithm is executed, which also increases the spatial variances.
On the contrary, a small $R$ may be used for datasets with high spatial heterogeneity, in which case fewer geographic objects should be taken into account.
Here, we take the synthetic dataset as an example to show how results change with the increasing $R$.
We perform our STICC algorithm with different $R$ ranging from 1 to 10.
As illustrated in Figure \ref{fig:radius}, both ARI and macro-F1 scores increase and reach the peak at $R=2$ or $R=3$, which means that taking several nearby geographic objects into account helps illustrate spatial dependencies within subregions; the two metrics then drop when $R$ is greater than 4. 
Thus, taking more nearby geographic objects into account may provide more complex spatial context information that could ``confuse" the model. 
In terms of the join count ratio, it is relatively stable when $R$ is less than 8, indicates that spatial contiguity of the observations is well preserved, and decreases when $R$ is greater than 9. Note that this pattern is only for our synthetic dataset. 

\begin{figure}[h]
    \centering
    \includegraphics[width=\textwidth]{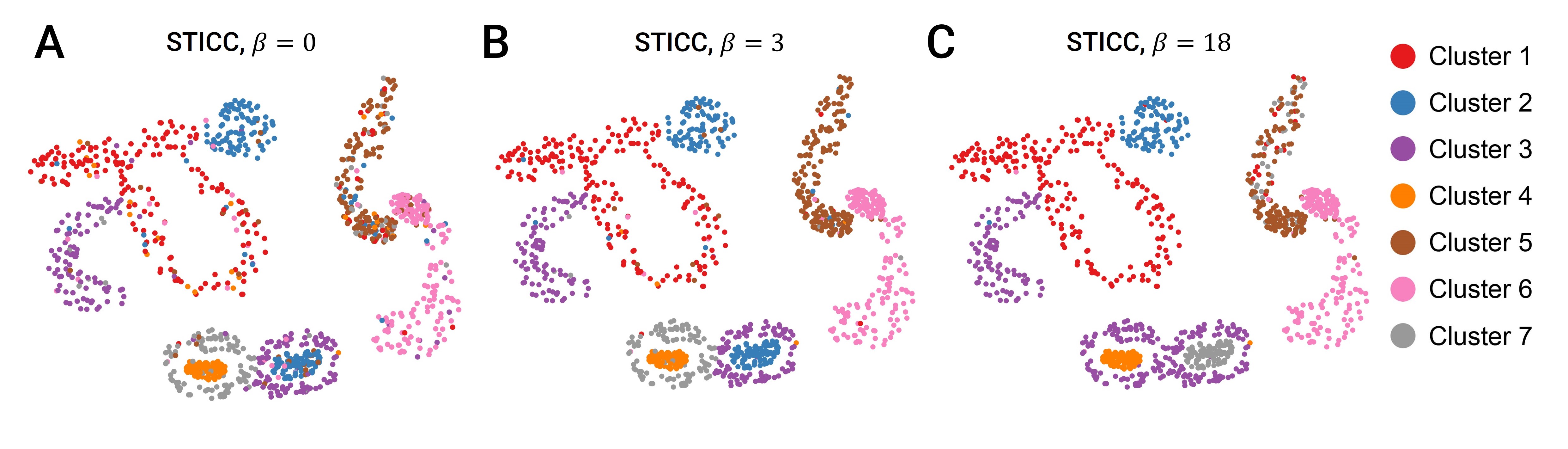}
    \caption{Results of the synthetic dataset using STICC with different $\beta$. (A) $\beta=0$, (B) $\beta=3$, (C) $\beta=18$.}
    \label{fig:beta}
\end{figure}

The $\beta$ controls the penalties of the spatial consistency strategy.
The larger the $\beta$, the more nearby points are grouped into the same cluster; while if $\beta=0$, no such strategy is carried out. 
An optimal $\beta$ might be inferred using parameter tuning.
Such a parameter impacts on the spatial contiguity maintained in the results.
Using the abovementioned synthetic dataset (in Figure \ref{fig:exp1_gt}), we provide an example here to show how clustering results change with different $\beta$. 
As illustrated in Figure \ref{fig:beta}, three sub-figures illustrate clustering results with $\beta=0, 3, 18$, respectively.
When $\beta$ is 0, no spatial consistency strategy is performed, and the points are relatively messy as points with different classes mixed up.
While with a relatively large $\beta=18$, nearby geographic objects are grouped into the same cluster, such as regions 7 and 9 (in purple) at the bottom, though they should belong to different clusters based on their attributes.

Recall that the hyperparameter $\lambda$ controls the level of sparsity on the inverse covariance matrices and can be used to prevent overfitting issues.
Although the inverse covariance matrix consists a $DR\times DR$ matrix in its general form, we restrict all the entries to a single value to save the hyperparameter search cost.
In practice it could be picked via cross-validation.

\subsection{Clustering Result Interpretation}
\begin{table}[h]
\caption{Betweenness centrality of attributes in different clusters of the synthetic dataset. The higher the betweenness centrality, the more ``important" the variable is in determining the cluster.
}
\begin{tabular}{@{}llllll@{}}
\toprule
          & Attribute A & Attribute B & Attribute C & Attribute D & Attribute E \\ \midrule
Cluster 1 & 0.00        & 0.00        & 0.50        & 0.83        & 0.00        \\
Cluster 2 & 0.00        & 0.00        & 0.50        & 0.83        & 0.25        \\
Cluster 3 & 0.83        & 0.00        & 0.08        & 0.83        & 0.58        \\
Cluster 4 & 0.42        & 0.00        & 0.25        & 0.58        & 0.92        \\
Cluster 5 & 0.17        & 1.00        & 0.00        & 0.42        & 0.67        \\
Cluster 6 & 0.83        & 0.42        & 0.17        & 0.58        & 0.00        \\
Cluster 7 & 0.00        & 0.17        & 0.92        & 0.33        & 0.83        \\ \bottomrule
\end{tabular}
\label{tab:interpret}
\end{table}
In this section, we discuss an approach for interpreting clustering characteristics. 
Since the structure of each cluster outputted from STICC is represented as a multilayer MRF network, network analysis approaches can be used for evaluating the properties of each cluster.
Betweenness centrality that assesses the number of shortest paths passing through the node has been frequently used as a measure of centrality in networks \citep{crucitti2006centrality}.
It emphasizes the importance of each node.
Taking the synthetic dataset as an example, the betweenness centrality score of each node in the output MRF network is computed and shown in Table \ref{tab:interpret}.
According to the table, the betweenness centrality of each attribute in each of the seven clusters is different. 
It can be inferred that attributes $C$ and $D$ play important roles in cluster 1, while attributes $A$, $D$ and $E$ are relatively important for cluster 3.
Though for the synthetic dataset, no physical meanings are attached to each variable, using network analysis to examine MRF networks provides possible ways in characterizing properties of each cluster.
In addition, not only the betweenness centrality might be used for cluster interpretation, but also other network measures, such as degree centrality and closeness centrality, might be utilized for different purposes.

\subsection{Implications for GeoAI and spatially explicit model}
The proposed STICC serves as an example of spatially explicit artificial intelligence techniques for geographic knowledge discovery \citep{goodchild2001issues,janowicz2020geoai}.
It shows how spatial thinking can be integrated into artificial intelligence, especially for the development of spatially explicit machine learning algorithms.
We made efforts to conceptualize ``space'' from the following two aspects.
On the one hand, by performing k-nearest neighbor analysis, a subregion is designed for each geographic object to model spatial dependencies among attributes using graph-based MRF.
On the other hand, a spatial consistency strategy is utilized by pairing the cluster of each geographic object to its nearest neighbor.
Such a strategy, as illustrated by the experiment results, is helpful in maintaining spatial contiguity.
If we take steps along these paths, more spatial and aspatial perspectives might be integrated into this algorithm.
For instance, in addition to \textit{distant} and \textit{near}, other spatial relationships such as \textit{direction} and \textit{scale} can be modeled into the algorithm \citep{zhu2019making,mai2018adcn}.
Spatial matrices such as Gaussian kernel, fixed bandwidth, and distance lags that express different spatial proxy relationships can be embedded into the STICC method as well \citep{yan2017itdl}.
In addition, recent studies regarding spatial clustering approaches mainly focus on two directions, either modifying existing algorithms to achieve better performances in classic spatial clustering tasks \citep{aydin2021quantitative, liu2019exploring}, or developing domain-specific algorithms for a specific field such as cartography \citep{wolf2021spatially}, human mobility \citep{liu2021snn_flow}, and geodemography \citep{grekousis2021local}, while limited attention was paid on the proposed RGPD problem.
We believe that this work is just a beginning for solving the RGPD problem. There are abundant spatial concepts that can be incorporated into the proposed algorithm so as to promote the state-of-the-art methods in GeoAI.

\subsection{Limitations and future work}
Although our proposed STICC method has shown promising results and outperformed other clustering methods in both case studies for solving the RGPD problem, we acknowledge the limitations of the proposed method, on which we discuss here.
As shown in Figure \ref{fig:exp1_result}, there are still some outliers in the results as several geographic objects are not labeled correctly. 
There are two potential reasons: one is that the selected parameters may not be appropriate; the other is that the algorithm may converge early before reaching the global optimum.
However, since all clustering approaches have pros and cons, our proposed STICC approach has potentials to solve a series of spatial clustering problems.

In addition, there are two potential directions that are worth exploration in the future work.
First, when constructing subregions for each geographic object, we only employ k-nearest neighbors in this research.
In fact, more approaches used for defining spatial adjacency relationships might be adapted in designing subregions. 
For example, a pre-defined distance radius can be used for the definition of subregions, i.e., geographic objects within a specific distance threshold are considered as neighborhoods of the center geographic object for constructing subregions.
Under such circumstance, spatial weights may be further assigned to nearby geographic objects according to inverse distance to construct a weighted neighborhood.
Second, given that the repeated geographic patterns are common, we believe that the proposed STICC might be appropriate for a variety of domain applications, including but not limited to climate type classification, semantic trajectory clustering, urban land use detection, remote sensing image segmentation, etc.
More experiments using this algorithm and its variations will be conducted in our future work.

\section{Conclusions}
In this paper, we develop a novel method to discover repeated clusters of multivariate geographic objects considering spatial contiguity.
The proposed STICC approach takes nearby geographic objects into account to construct subregions rather than treating each object in isolation.
Markov random fields are used for representing dependencies among attributes inside subregions.
Then, a spatial consistency strategy is used for forcing nearby geographic objects to be assigned into the same group.
An adapted E-M strategy is used for cluster assignment and parameter updates.
Two case studies including a synthetic dataset and a set of POI category identification tasks prove that our proposed method has yielded good results in solving the RGPD problem in comparison with other attribute-based, density-based, or regionalization-based clustering methods.
It could discover repeated clusters accurately while maintain spatial contiguity at the same time, which is more in line with the natural distribution pattern of geographical objects.
Recall the Figure \ref{fig:framework}, the STICC method achieves the balance between repeated patterns and spatial contiguity.
This new spatial clustering approach can support many other applications in geography, remote sensing, transportation, urban planning, and social sciences.

\section*{Data and codes availability statement}
The data and codes that support the findings of this study are available in figshare.com with the link 
\url{https://doi.org/10.6084/m9.figshare.15170898.v1} and on the Github repository \url{https://github.com/GeoDS/STICC}.


\section*{Acknowledgement}
Yuhao Kang acknowledges the support by the Trewartha Research Award, Department of the Geography, University of Wisconsin-Madison. Song Gao and Jinmeng Rao acknowledge the support by the American Family Insurance Data Science Institute at the University of Wisconsin-Madison and the National Science Foundation funded AI institute (Grant No.2112606) for Intelligent Cyberinfrastructure with Computational Learning in the Environment (ICICLE).
Fan Zhang would like to thank the support by the National Natural Science Foundation of China under Grant 41901321. Ignavier Ng would like to acknowledge the support by the National Institutes of Health (NIH) under Contract R01HL159805.
Any opinions, findings, and conclusions or recommendations expressed in this material are those of the author(s) and do not necessarily reflect the views of the funders.

\section*{Notes on contributors}
\noindent \textbf{Yuhao Kang} is a Ph.D. student in GIScience at the Department of Geography, University of Wisconsin-Madison. He holds a B.S. degree in Geographic Information Science at Wuhan University. His main research interests include Place-Based GIS, GeoAI, and Social Sensing.

\noindent \textbf{Kunlin Wu} is a graduate student in the School of Resource and Environmental Sciences at Wuhan University. His research interests include remote sensing, spatial data mining and auralization of spatial data.

\noindent \textbf{Song Gao} is an Assistant Professor in GIScience at the Department of Geography, University of Wisconsin-Madison. He holds a Ph.D. in Geography at the University of California, Santa Barbara. His main research interests include Place-Based GIS, Geospatial Data Science, and GeoAI approaches to Human Mobility and Social Sensing.

\noindent \textbf{Ignavier Ng} is a Ph.D. student at Carnegie Mellon University. His research interests include causal discovery and machine learning.

\noindent \textbf{Jinmeng Rao} is a Ph.D. student at the Department of Geography, University of Wisconsin-Madison. His research interests include GeoAI, Privacy-Preserving AI, and Location Privacy.

\noindent \textbf{Shan Ye} is a Ph.D. candidate at the Department of Geoscience, University of Wisconsin-Madison. His research focuses on quantitative stratigraphy and paleoclimate data science.

\noindent \textbf{Fan Zhang} is a Senior Research Associate at Senseable City Lab, Massachusetts Institute of Technology. His research interests include Urban Data Science, Visual Intelligence, GeoAI, and Social Sensing.

\noindent \textbf{Teng Fei} is currently an Associate Professor in the School of Resource and Environmental Sciences at Wuhan University. His research focuses on remote sensing, urban data analysis, social sensing and ecological modelling.

\bibliographystyle{apalike}
\bibliography{ref}

\begin{thebibliography}{}

\bibitem[Akaike, 1973]{akaike1973block}
Akaike, H. (1973).
\newblock Block toeplitz matrix inversion.
\newblock {\em SIAM Journal on Applied Mathematics}, 24(2):234--241.

\bibitem[Aldstadt, 2010]{aldstadt2010spatial}
Aldstadt, J. (2010).
\newblock Spatial clustering.
\newblock In {\em Handbook of applied spatial analysis}, pages 279--300.
  Springer.

\bibitem[Anderberg, 1973]{anderberg1973cluster}
Anderberg, M.~R. (1973).
\newblock {\em Cluster analysis for applications}, volume~19.
\newblock Academic Press, New York.

\bibitem[Ankerst et~al., 1999]{ankerst1999optics}
Ankerst, M., Breunig, M.~M., Kriegel, H.-P., and Sander, J. (1999).
\newblock Optics: Ordering points to identify the clustering structure.
\newblock {\em ACM Sigmod record}, 28(2):49--60.

\bibitem[Aragam and Zhou, 2015]{aragam2015concave}
Aragam, B. and Zhou, Q. (2015).
\newblock Concave penalized estimation of sparse gaussian {Bayesian} networks.
\newblock {\em Journal of Machine Learning Research}, 16(1):2273--2328.

\bibitem[Assun{\c{c}}{\~a}o et~al., 2006]{assunccao2006efficient}
Assun{\c{c}}{\~a}o, R.~M., Neves, M.~C., C{\^a}mara, G., and da~Costa~Freitas,
  C. (2006).
\newblock Efficient regionalization techniques for socio-economic geographical
  units using minimum spanning trees.
\newblock {\em International Journal of Geographical Information Science},
  20(7):797--811.

\bibitem[Aydin et~al., 2018]{aydin2018skater}
Aydin, O., Janikas, M.~V., Assun{\c{c}}{\~a}o, R., and Lee, T.-H. (2018).
\newblock Skater-con: Unsupervised regionalization via stochastic tree
  partitioning within a consensus framework using random spanning trees.
\newblock In {\em Proceedings of the 2nd ACM SIGSPATIAL International Workshop
  on AI for Geographic Knowledge Discovery}, pages 33--42.

\bibitem[Aydin et~al., 2021]{aydin2021quantitative}
Aydin, O., Janikas, M.~V., Assun{\c{c}}{\~a}o, R.~M., and Lee, T.-H. (2021).
\newblock A quantitative comparison of regionalization methods.
\newblock {\em International Journal of Geographical Information Science},
  pages 1--29.

\bibitem[Ba{\c{c}}{\~a}o et~al., 2005]{baccao2005self}
Ba{\c{c}}{\~a}o, F., Lobo, V., and Painho, M. (2005).
\newblock Self-organizing maps as substitutes for k-means clustering.
\newblock In {\em International Conference on Computational Science}, pages
  476--483. Springer.

\bibitem[Boyd et~al., 2011]{boyd2011admm}
Boyd, S., Parikh, N., Chu, E., Peleato, B., and Eckstein, J. (2011).
\newblock Distributed optimization and statistical learning via the alternating
  direction method of multipliers.
\newblock {\em Foundations and Trends in Machine Learning}, 3(1):1--122.

\bibitem[Candes and Tao, 2005]{candes2005decoding}
Candes, E.~J. and Tao, T. (2005).
\newblock Decoding by linear programming.
\newblock {\em IEEE Transactions on Information Theory}, 51(12):4203--4215.

\bibitem[Chen et~al., 2018]{chen2018hispatialcluster}
Chen, Y., Huang, Z., Pei, T., and Liu, Y. (2018).
\newblock Hispatialcluster: A novel high-performance software tool for
  clustering massive spatial points.
\newblock {\em Transactions in GIS}, 22(5):1275--1298.

\bibitem[Cliff and Ord, 1973]{cliff1973spatial}
Cliff, A.~D. and Ord, J.~K. (1973).
\newblock {\em Spatial autocorrelation}.
\newblock Pion.

\bibitem[Crucitti et~al., 2006]{crucitti2006centrality}
Crucitti, P., Latora, V., and Porta, S. (2006).
\newblock Centrality measures in spatial networks of urban streets.
\newblock {\em Physical Review E}, 73(3):036125.

\bibitem[Dacey, 1965]{dacey1965review}
Dacey, M.~F. (1965).
\newblock A review on measures of contiguity for two and k-color maps.
\newblock Technical report, NORTHWESTERN UNIV EVANSTON ILL.

\bibitem[Deng et~al., 2011]{deng2011adaptive}
Deng, M., Liu, Q., Cheng, T., and Shi, Y. (2011).
\newblock An adaptive spatial clustering algorithm based on delaunay
  triangulation.
\newblock {\em Computers, Environment and Urban Systems}, 35(4):320--332.

\bibitem[Deng et~al., 2019]{deng2019density}
Deng, M., Yang, X., Shi, Y., Gong, J., Liu, Y., and Liu, H. (2019).
\newblock A density-based approach for detecting network-constrained clusters
  in spatial point events.
\newblock {\em International Journal of Geographical Information Science},
  33(3):466--488.

\bibitem[Donoho, 2006]{donoho2006compressed}
Donoho, D.~L. (2006).
\newblock Compressed sensing.
\newblock {\em IEEE Transactions on Information Theory}, 52(4):1289--1306.

\bibitem[Duque et~al., 2011]{duque2011p}
Duque, J.~C., Church, R.~L., and Middleton, R.~S. (2011).
\newblock The p-regions problem.
\newblock {\em Geographical Analysis}, 43(1):104--126.

\bibitem[Duque et~al., 2007]{duque2007supervised}
Duque, J.~C., Ramos, R., and Suri{\~n}ach, J. (2007).
\newblock Supervised regionalization methods: A survey.
\newblock {\em International Regional Science Review}, 30(3):195--220.

\bibitem[Esri, 2021]{esri2021spatial}
Esri (2021).
\newblock {How Spatially Constrained Multivariate Clustering works}.
\newblock
  \url{https://pro.arcgis.com/en/pro-app/latest/tool-reference/spatial-statistics/how-spatially-constrained-multivariate-clustering-works.htm}.
\newblock [Online; accessed 24-June-2021].

\bibitem[Ester et~al., 1996]{ester1996density}
Ester, M., Kriegel, H.-P., Sander, J., Xu, X., et~al. (1996).
\newblock A density-based algorithm for discovering clusters in large spatial
  databases with noise.
\newblock In {\em Int. Conf. Knowledge Discovery and Data Mining}, volume~96,
  pages 226--231.

\bibitem[Estivill-Castro and Lee, 2002]{estivill2002argument}
Estivill-Castro, V. and Lee, I. (2002).
\newblock Argument free clustering for large spatial point-data sets via
  boundary extraction from delaunay diagram.
\newblock {\em Computers, Environment and urban systems}, 26(4):315--334.

\bibitem[Friedman et~al., 2008]{friedman2008sparse}
Friedman, J., Hastie, T., and Tibshirani, R. (2008).
\newblock Sparse inverse covariance estimation with the graphical {Lasso}.
\newblock {\em Biostatistics}, 9:432--41.

\bibitem[Fujino et~al., 2008]{fujino2008multi}
Fujino, A., Isozaki, H., and Suzuki, J. (2008).
\newblock Multi-label text categorization with model combination based on
  f1-score maximization.
\newblock In {\em Proceedings of the Third International Joint Conference on
  Natural Language Processing: Volume-II}.

\bibitem[Gao et~al., 2017]{gao2017extracting}
Gao, S., Janowicz, K., and Couclelis, H. (2017).
\newblock Extracting urban functional regions from points of interest and human
  activities on location-based social networks.
\newblock {\em Transactions in GIS}, 21(3):446--467.

\bibitem[Goodchild, 2001]{goodchild2001issues}
Goodchild, M. (2001).
\newblock Issues in spatially explicit modeling.
\newblock {\em Agent-based models of land-use and land-cover change}, pages
  13--17.

\bibitem[Goodchild, 2004]{goodchild2004giscience}
Goodchild, M.~F. (2004).
\newblock Giscience, geography, form, and process.
\newblock {\em Annals of the Association of American Geographers},
  94(4):709--714.

\bibitem[Grekousis, 2021]{grekousis2021local}
Grekousis, G. (2021).
\newblock Local fuzzy geographically weighted clustering: a new method for
  geodemographic segmentation.
\newblock {\em International Journal of Geographical Information Science},
  35(1):152--174.

\bibitem[Guha et~al., 1998]{guha1998cure}
Guha, S., Rastogi, R., and Shim, K. (1998).
\newblock Cure: An efficient clustering algorithm for large databases.
\newblock {\em ACM Sigmod record}, 27(2):73--84.

\bibitem[Hallac et~al., 2017]{hallac2017toeplitz}
Hallac, D., Vare, S., Boyd, S., and Leskovec, J. (2017).
\newblock Toeplitz inverse covariance-based clustering of multivariate time
  series data.
\newblock In {\em Proceedings of the 23rd ACM SIGKDD International Conference
  on Knowledge Discovery and Data Mining}, pages 215--223.

\bibitem[Hastie et~al., 2015]{hastie2015statistical}
Hastie, T., Tibshirani, R., and Wainwright, M. (2015).
\newblock {\em Statistical Learning with Sparsity: The Lasso and
  Generalizations}.
\newblock Chapman and Hall/CRC.

\bibitem[Henriques et~al., 2009]{henriques2009geosom}
Henriques, R., Ba{\c{c}}{\~a}o, F., and Lobo, V. (2009).
\newblock Geosom suite: a tool for spatial clustering.
\newblock In {\em International Conference on Computational Science and Its
  Applications}, pages 453--466. Springer.

\bibitem[Hinneburg et~al., 1998]{hinneburg1998efficient}
Hinneburg, A., Keim, D.~A., et~al. (1998).
\newblock An efficient approach to clustering in large multimedia databases
  with noise.
\newblock In {\em KDD}, volume~98, pages 58--65.

\bibitem[Hsieh et~al., 2014]{hsieh2014quic}
Hsieh, C.-J., Sustik, M.~A., Dhillon, I.~S., and Ravikumar, P. (2014).
\newblock {QUIC}: Quadratic approximation for sparse inverse covariance
  estimation.
\newblock {\em Journal of Machine Learning Research}, 15(83):2911--2947.

\bibitem[Hu et~al., 2015]{hu2015extracting}
Hu, Y., Gao, S., Janowicz, K., Yu, B., Li, W., and Prasad, S. (2015).
\newblock Extracting and understanding urban areas of interest using geotagged
  photos.
\newblock {\em Computers, Environment and Urban Systems}, 54:240--254.

\bibitem[Janowicz et~al., 2020]{janowicz2020geoai}
Janowicz, K., Gao, S., McKenzie, G., Hu, Y., and Bhaduri, B. (2020).
\newblock {GeoAI}: spatially explicit artificial intelligence techniques for
  geographic knowledge discovery and beyond.
\newblock 34(4):625--636.

\bibitem[Kang et~al., 2019]{kang2019extracting}
Kang, Y., Jia, Q., Gao, S., Zeng, X., Wang, Y., Angsuesser, S., Liu, Y., Ye,
  X., and Fei, T. (2019).
\newblock Extracting human emotions at different places based on facial
  expressions and spatial clustering analysis.
\newblock {\em Transactions in GIS}, 23(3):450--480.

\bibitem[Kodinariya and Makwana, 2013]{kodinariya2013review}
Kodinariya, T.~M. and Makwana, P.~R. (2013).
\newblock Review on determining number of cluster in k-means clustering.
\newblock {\em International Journal}, 1(6):90--95.

\bibitem[Koller and Friedman, 2009]{koller09probabilistic}
Koller, D. and Friedman, N. (2009).
\newblock {\em Probabilistic Graphical Models: Principles and Techniques}.
\newblock MIT Press, Cambridge, MA.

\bibitem[Liqiang et~al., 2013]{liqiang2013spatial}
Liqiang, Z., Hao, D., Dong, C., and Zhen, W. (2013).
\newblock A spatial cognition-based urban building clustering approach and its
  applications.
\newblock {\em International Journal of Geographical Information Science},
  27(4):721--740.

\bibitem[Liu et~al., 2020]{liu2020investigating}
Liu, K., Qiu, P., Gao, S., Lu, F., Jiang, J., and Yin, L. (2020).
\newblock Investigating urban metro stations as cognitive places in cities
  using points of interest.
\newblock {\em Cities}, 97:102561.

\bibitem[Liu et~al., 2012]{liu2012density}
Liu, Q., Deng, M., Shi, Y., and Wang, J. (2012).
\newblock A density-based spatial clustering algorithm considering both spatial
  proximity and attribute similarity.
\newblock {\em Computers \& Geosciences}, 46:296--309.

\bibitem[Liu et~al., 2021]{liu2021snn_flow}
Liu, Q., Yang, J., Deng, M., Song, C., and Liu, W. (2021).
\newblock Snn\_flow: a shared nearest-neighbor-based clustering method for
  inhomogeneous origin-destination flows.
\newblock {\em International Journal of Geographical Information Science},
  pages 1--27.

\bibitem[Liu et~al., 2019]{liu2019exploring}
Liu, X., Huang, Q., and Gao, S. (2019).
\newblock Exploring the uncertainty of activity zone detection using digital
  footprints with multi-scaled dbscan.
\newblock {\em International Journal of Geographical Information Science},
  33(6):1196--1223.

\bibitem[MacQueen et~al., 1967]{macqueen1967some}
MacQueen, J. et~al. (1967).
\newblock Some methods for classification and analysis of multivariate
  observations.
\newblock In {\em Proceedings of the fifth Berkeley symposium on mathematical
  statistics and probability}, volume~1, pages 281--297. Oakland, CA, USA.

\bibitem[Mai et~al., 2018]{mai2018adcn}
Mai, G., Janowicz, K., Hu, Y., and Gao, S. (2018).
\newblock Adcn: An anisotropic density-based clustering algorithm for
  discovering spatial point patterns with noise.
\newblock {\em Transactions in GIS}, 22(1):348--369.

\bibitem[Miller and Han, 2009]{miller2009geographic}
Miller, H.~J. and Han, J. (2009).
\newblock {\em Geographic data mining and knowledge discovery}.
\newblock CRC press.

\bibitem[Moayedi et~al., 2019]{moayedi2019evaluation}
Moayedi, A., Abbaspour, R.~A., and Chehreghan, A. (2019).
\newblock An evaluation of the efficiency of similarity functions in
  density-based clustering of spatial trajectories.
\newblock {\em Annals of GIS}, 25(4):313--327.

\bibitem[Murray and Estivill-Castro, 1998]{murray1998cluster}
Murray, A.~T. and Estivill-Castro, V. (1998).
\newblock Cluster discovery techniques for exploratory spatial data analysis.
\newblock {\em International journal of geographical information science},
  12(5):431--443.

\bibitem[Murray and Shyy, 2000]{murray2000integrating}
Murray, A.~T. and Shyy, T.-K. (2000).
\newblock Integrating attribute and space characteristics in choropleth display
  and spatial data mining.
\newblock {\em International Journal of Geographical Information Science},
  14(7):649--667.

\bibitem[Ng et~al., 2020]{ng2020role}
Ng, I., Ghassami, A., and Zhang, K. (2020).
\newblock On the role of sparsity and {DAG} constraints for learning linear
  {DAGs}.
\newblock In {\em Advances in Neural Information Processing Systems}.

\bibitem[Nosovskiy et~al., 2008]{nosovskiy2008automatic}
Nosovskiy, G.~V., Liu, D., and Sourina, O. (2008).
\newblock Automatic clustering and boundary detection algorithm based on
  adaptive influence function.
\newblock {\em Pattern Recognition}, 41(9):2757--2776.

\bibitem[Novikov, 2019]{novikov2019pyclustering}
Novikov, A.~V. (2019).
\newblock Pyclustering: Data mining library.
\newblock {\em Journal of Open Source Software}, 4(36):1230.

\bibitem[Ogbuabor and Ugwoke, 2018]{ogbuabor2018clustering}
Ogbuabor, G. and Ugwoke, F. (2018).
\newblock Clustering algorithm for a healthcare dataset using silhouette score
  value.
\newblock {\em International Journal of Computer Science \& Information
  Technology (IJCSIT)}, 10(2):27--37.

\bibitem[Pei et~al., 2015]{pei2015density}
Pei, T., Wang, W., Zhang, H., Ma, T., Du, Y., and Zhou, C. (2015).
\newblock Density-based clustering for data containing two types of points.
\newblock {\em International Journal of Geographical Information Science},
  29(2):175--193.

\bibitem[Pei et~al., 2006]{pei2006new}
Pei, T., Zhu, A.-X., Zhou, C., Li, B., and Qin, C. (2006).
\newblock A new approach to the nearest-neighbour method to discover cluster
  features in overlaid spatial point processes.
\newblock {\em International Journal of Geographical Information Science},
  20(2):153--168.

\bibitem[Perruchet, 1983]{perruchet1983constrained}
Perruchet, C. (1983).
\newblock Constrained agglomerative hierarchical classification.
\newblock {\em Pattern Recognition}, 16(2):213--217.

\bibitem[Rao et~al., 2020]{rao2020lstm}
Rao, J., Gao, S., Kang, Y., and Huang, Q. (2020).
\newblock Lstm-trajgan: A deep learning approach to trajectory privacy
  protection.
\newblock {\em arXiv preprint arXiv:2006.10521}.

\bibitem[Rasmussen et~al., 1999]{rasmussen1999infinite}
Rasmussen, C.~E. et~al. (1999).
\newblock The infinite gaussian mixture model.
\newblock In {\em NIPS}, volume~12, pages 554--560.

\bibitem[Rey and Anselin, 2010]{rey2010pysal}
Rey, S.~J. and Anselin, L. (2010).
\newblock Pysal: A python library of spatial analytical methods.
\newblock In {\em Handbook of applied spatial analysis}, pages 175--193.
  Springer.

\bibitem[Rue and Held, 2005]{rue2005gaussian}
Rue, H. and Held, L. (2005).
\newblock {\em Gaussian Markov Random Fields: Theory And Applications
  (Monographs on Statistics and Applied Probability)}.
\newblock Chapman Hall/CRC.

\bibitem[Scheinberg et~al., 2010]{scheinberg2010sparse}
Scheinberg, K., Ma, S., and Goldfarb, D. (2010).
\newblock Sparse inverse covariance selection via alternating linearization
  methods.
\newblock In {\em Advances in Neural Information Processing Systems},
  volume~23.

\bibitem[Soliman et~al., 2017]{soliman2017social}
Soliman, A., Soltani, K., Yin, J., Padmanabhan, A., and Wang, S. (2017).
\newblock Social sensing of urban land use based on analysis of twitter
  users’ mobility patterns.
\newblock {\em PloS one}, 12(7):e0181657.

\bibitem[Steinley, 2004]{steinley2004properties}
Steinley, D. (2004).
\newblock Properties of the hubert-arable adjusted rand index.
\newblock {\em Psychological methods}, 9(3):386.

\bibitem[Sugar and James, 2003]{sugar2003finding}
Sugar, C.~A. and James, G.~M. (2003).
\newblock Finding the number of clusters in a dataset: An information-theoretic
  approach.
\newblock {\em Journal of the American Statistical Association},
  98(463):750--763.

\bibitem[Syakur et~al., 2018]{syakur2018integration}
Syakur, M., Khotimah, B., Rochman, E., and Satoto, B.~D. (2018).
\newblock Integration k-means clustering method and elbow method for
  identification of the best customer profile cluster.
\newblock In {\em IOP Conference Series: Materials Science and Engineering},
  volume 336, page 012017. IOP Publishing.

\bibitem[Tibshirani, 1996]{tibshirani1996lasso}
Tibshirani, R. (1996).
\newblock Regression shrinkage and selection via the lasso.
\newblock {\em Journal of the Royal Statistical Society. Series B
  (Methodological)}, 58(1):267--288.

\bibitem[Tobler, 1970]{tobler1970computer}
Tobler, W.~R. (1970).
\newblock A computer movie simulating urban growth in the detroit region.
\newblock {\em Economic geography}, 46(sup1):234--240.

\bibitem[Viterbi, 1967]{viterbi1967error}
Viterbi, A.~J. (1967).
\newblock Error bounds for convolutional codes and an asymptotically optimum
  decoding algorithm.
\newblock {\em IEEE Transactions on Information Theory}, 13(2):260--269.

\bibitem[Wang, 2020]{wang2020public}
Wang, F. (2020).
\newblock Why public health needs gis: A methodological overview.
\newblock {\em Annals of GIS}, 26(1):1--12.

\bibitem[Webster and Burrough, 1972]{webster1972computer}
Webster, R. and Burrough, P. (1972).
\newblock Computer-based soil mapping of small areas from sample data: I.
  multivariate classification and ordination.
\newblock {\em Journal of Soil Science}, 23(2):210--221.

\bibitem[Wilson et~al., 2020]{wilson2020five}
Wilson, J.~P., Butler, K., Gao, S., Hu, Y., Li, W., and Wright, D.~J. (2020).
\newblock A five-star guide for achieving replicability and reproducibility
  when working with gis software and algorithms.
\newblock {\em Annals of the American Association of Geographers}, pages 1--7.

\bibitem[Wolf, 2021]{wolf2021spatially}
Wolf, L.~J. (2021).
\newblock Spatially--encouraged spectral clustering: a technique for blending
  map typologies and regionalization.
\newblock {\em International Journal of Geographical Information Science},
  35(11):2356--2373.

\bibitem[Wu et~al., 2014]{wu2014intra}
Wu, L., Zhi, Y., Sui, Z., and Liu, Y. (2014).
\newblock Intra-urban human mobility and activity transition: Evidence from
  social media check-in data.
\newblock {\em PloS one}, 9(5):e97010.

\bibitem[Xing and Meng, 2018]{xing2018integrating}
Xing, H. and Meng, Y. (2018).
\newblock Integrating landscape metrics and socioeconomic features for urban
  functional region classification.
\newblock {\em Computers, Environment and Urban Systems}, 72:134--145.

\bibitem[Yan et~al., 2017]{yan2017itdl}
Yan, B., Janowicz, K., Mai, G., and Gao, S. (2017).
\newblock From itdl to place2vec: Reasoning about place type similarity and
  relatedness by learning embeddings from augmented spatial contexts.
\newblock In {\em Proceedings of the 25th ACM SIGSPATIAL international
  conference on advances in geographic information systems}, pages 1--10.

\bibitem[Yang et~al., 2014]{yang2014modeling}
Yang, D., Zhang, D., Zheng, V.~W., and Yu, Z. (2014).
\newblock Modeling user activity preference by leveraging user spatial temporal
  characteristics in lbsns.
\newblock {\em IEEE Transactions on Systems, Man, and Cybernetics: Systems},
  45(1):129--142.

\bibitem[Yuan et~al., 2014]{yuan2014discovering}
Yuan, N.~J., Zheng, Y., Xie, X., Wang, Y., Zheng, K., and Xiong, H. (2014).
\newblock Discovering urban functional zones using latent activity
  trajectories.
\newblock {\em IEEE Transactions on Knowledge and Data Engineering},
  27(3):712--725.

\bibitem[Zhang et~al., 1996]{zhang1996birch}
Zhang, T., Ramakrishnan, R., and Livny, M. (1996).
\newblock Birch: an efficient data clustering method for very large databases.
\newblock {\em ACM sigmod record}, 25(2):103--114.

\bibitem[Zhao et~al., 2016]{zhao2016understanding}
Zhao, Z., Shaw, S.-L., Xu, Y., Lu, F., Chen, J., and Yin, L. (2016).
\newblock Understanding the bias of call detail records in human mobility
  research.
\newblock {\em International Journal of Geographical Information Science},
  30(9):1738--1762.

\bibitem[Zhu et~al., 2018]{zhu2018spatial}
Zhu, A.-X., Lu, G., Liu, J., Qin, C.-Z., and Zhou, C. (2018).
\newblock Spatial prediction based on third law of geography.
\newblock {\em Annals of GIS}, 24(4):225--240.

\bibitem[Zhu et~al., 2019]{zhu2019making}
Zhu, R., Janowicz, K., and Mai, G. (2019).
\newblock Making direction a first-class citizen of tobler's first law of
  geography.
\newblock {\em Transactions in GIS}, 23(3):398--416.

\bibitem[Zou, 2006]{zou2006adaptive}
Zou, H. (2006).
\newblock The adaptive {Lasso} and its oracle properties.
\newblock {\em Journal of the American Statistical Association},
  101(476):1418--1429.

\end{thebibliography}
\end{document}